\pdfoutput=1

\documentclass[11pt]{article}

\usepackage[]{acl}

\usepackage{times}
\usepackage{latexsym}
\usepackage{multirow}
\usepackage[T1]{fontenc}

\usepackage[utf8]{inputenc}

\usepackage{microtype}

\usepackage{inconsolata}

\usepackage{enumitem}

\usepackage{graphicx}
\usepackage{paralist}
\usepackage{amsmath}
\usepackage{booktabs} 
\usepackage{pifont}
\usepackage{soul}

\sethlcolor{yellow}

\usepackage{marvosym}

\title{M$^3$AV: A Multimodal, Multigenre, and Multipurpose Audio-Visual Academic Lecture Dataset}

\author{
    Zhe~Chen$^{1}$,
    Heyang~Liu$^{1}$,
    Wenyi~Yu$^{2}$, 
    Guangzhi~Sun$^{3}$, 
    Hongcheng~Liu$^{1}$
    \\ 
    \bf  
    Ji~Wu$^{2}$, 
    Chao~Zhang$^{2}$\textsuperscript{\Letter}, 
    Yu~Wang$^{1,4}$\textsuperscript{\Letter},
    Yanfeng~Wang$^{1,4}$ \vspace{2mm}
    \\ 
     \small
    \begin{tabular}{c} 
    $^1$Cooperative Medianet Innovation Center, Shanghai JiaoTong University \\
    $^2$Department of Electronic Engineering, Tsinghua University \\
    $^3$University of Cambridge Department of Engineering~~~~$^4$Shanghai AI Laboratory \vspace{2mm}\\
    \end{tabular}
    \\ 
    \small
    \begin{tabular}{c}
    \texttt{\{chenzhe2018,liuheyang,hongcheng\_liu,yuwangsjtu,wangyanfeng622\}@sjtu.edu.cn} \\
    \texttt{ywy22@mails.tsinghua.edu.cn, gs534@cam.ac.uk, \{wuji\_ee,cz277\}@tsinghua.edu.cn} \\
    \end{tabular}
}

\begin{document}
\maketitle
\renewcommand{\thefootnote}{}
\footnotetext{\Letter: Corresponding author.}
\renewcommand{\thefootnote}{\arabic{footnote}} 

\begin{abstract}
Publishing open-source academic video recordings is an emergent and prevalent approach to sharing knowledge online. Such videos carry rich multimodal information including speech, the facial and body movements of the speakers, as well as the texts and pictures in the slides and possibly even the papers. 
Although multiple academic video datasets have been constructed and released, few of them support both multimodal content recognition and understanding tasks, which is partially due to the lack of high-quality human annotations. In this paper, we propose a novel multimodal, multigenre, and multipurpose audio-visual academic lecture dataset (M$^3$AV), which has almost 367 hours of videos from five sources covering computer science, mathematics, and medical and biology topics. With high-quality human annotations of the slide text and spoken words, in particular high-valued name entities, the dataset can be used for multiple audio-visual recognition and understanding tasks. Evaluations performed on contextual speech recognition, speech synthesis, and slide and script generation tasks demonstrate that the diversity of M$^3$AV makes it a challenging dataset\footnote{Project website: \url{https://jack-zc8.github.io/M3AV-dataset-page}}.

\end{abstract}
\section{Introduction}

The rapid progress of technology has brought numerous academic presentations and talks available on the web with open access~\cite{Lev2019, Atri2021, lee2022multimodal, Lee2023}. These resources are particularly helpful to researchers as they contain rich specialised knowledge in auditory and visual modalities. With the development of neural-based AI systems, it is desired that AI could comprehend and process these multimodal information resources to assist scientists in accelerating their research. Abilities, including transcribing and generating speech for a presentation with the aid of slides, as well as presentation slide or script generation, are particularly useful for researchers to conduct investigations and prepare presentations.

\begin{figure*}[htbp]
    \centering
    \includegraphics[width=1\linewidth]{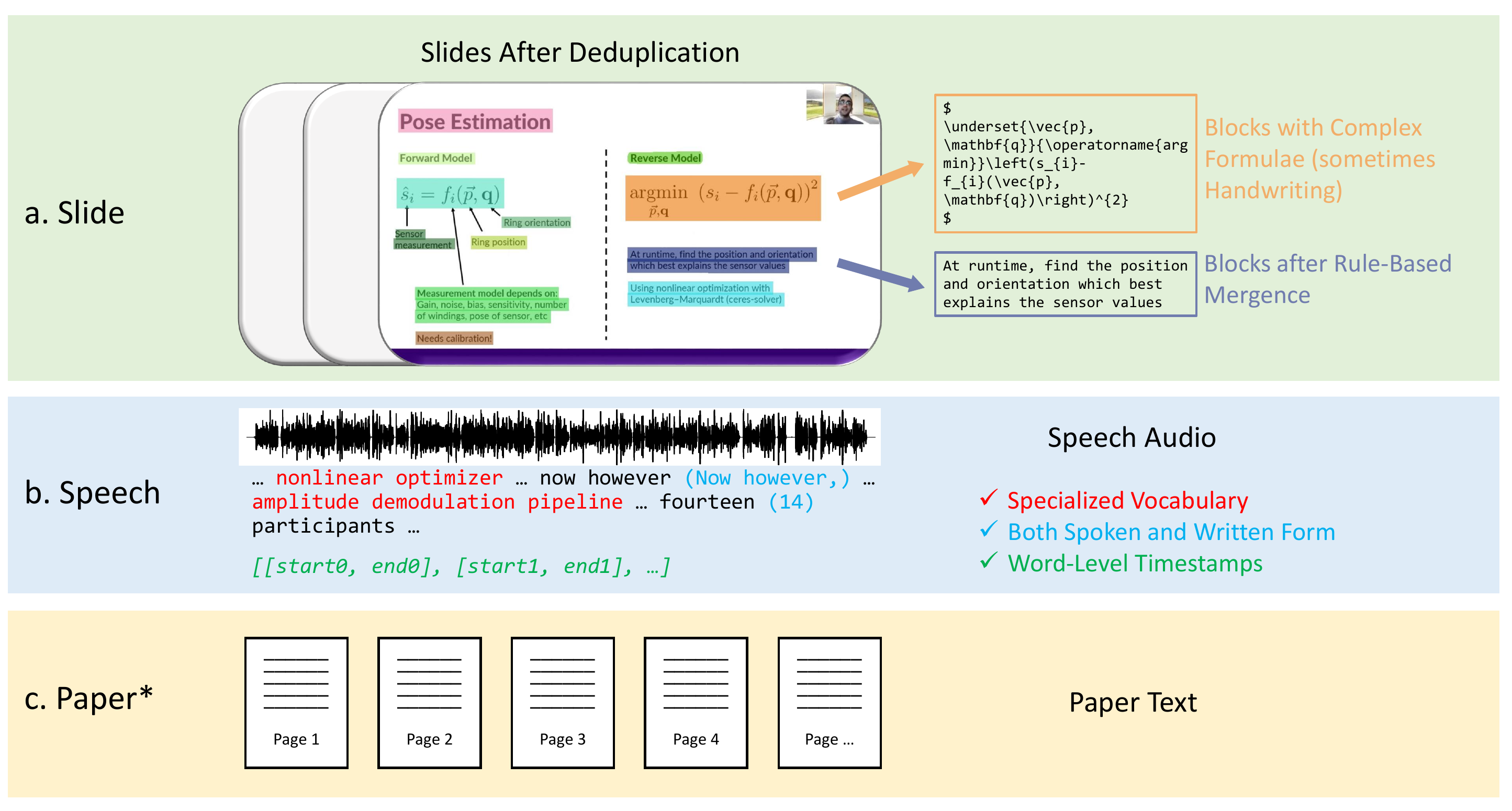}
    \caption{The overview of our M$^3$AV dataset. The first component is slides annotated with simple and complex blocks. They will be merged following some rules. The second component is speech containing special vocabulary, spoken and written forms, and word-level timestamps. The third component is the paper corresponding to the video. The asterisk (*) denotes that only computer science videos have corresponding papers.}
    \label{fig:dataset}
\end{figure*}

As a natural data source containing multimodal content and knowledge, academic lectures have been explored to construct a series of datasets. Some datasets focus on the evaluation of the model's ability to recognize multimodal content~\cite{Dutta2018, Wang2023}, others focus on promoting models to understand multimodal knowledge in academic videos~\cite{Lev2019, Li2019, Atri2021, lee2022multimodal, Lee2023}.
However, none of them has yet simultaneously paid attention to the model's ability to recognize multimodal content and understand rich academic knowledge, which is partly due to their respective lack of adequate manual labelling. The homogeneity of holding two characteristics is critical to the implementation of end-to-end academic recognition and understanding systems.

To fill the gap, we release the Multimodal, Multigenre, and Multipurpose Audio-Visual Academic Lecture Dataset (M$^3$AV) as shown in Figure~\ref{fig:dataset}. The dataset contains selected academic lectures and presentation videos from multiple fields, including computer science, biomedical science, and mathematics. Each video is endowed with highly qualified speech transcriptions and OCR labels for both printed and handwritten characters, including complex mathematical formulae.
In addition, academic papers for some videos are also provided to supplement knowledge more extensively. 

Together with the dataset, three benchmark tasks are also proposed that reflect the perception and understanding of the multimodal information in those videos. This includes automatic speech recognition (ASR) with a particular focus on contextual ASR (CASR), spontaneous text-to-speech (TTS) synthesis, and slide and script generation (SSG). Representative baseline models are used to benchmark the performance for each task. Through our experiments, it is observed that the existing models showcase limited performance in perceiving and understanding the multimodal content in this dataset, and the rich academic knowledge is not utilized effectively.

\section{Related Work}

\begin{table*}[htbp]
\caption{Comparison with other academic lecture-based datasets in terms of data types and designed tasks. ``A'' denotes fully automated processing and ``M'' denotes fully or partially manual labelling.}
\vspace{-0.2cm}
\centering
\resizebox{1\textwidth}{!}{
\begin{tabular}{l|ccccc|c|c|cc}
\toprule
 & \multicolumn{5}{c|}{\bf Slide Data} & \multicolumn{1}{c|}{\bf Speech Data} & \multicolumn{1}{c|}{\bf Paper Data} & \multicolumn{2}{c}{\bf Dataset Tasks} \\ \midrule
 
 & \multicolumn{1}{c|}{Segments} & \multicolumn{1}{c|}{Figures} & \multicolumn{1}{c|}{Text} & \multicolumn{1}{c|}{Handwriting} & \multicolumn{1}{c|}{Formulae} & \multicolumn{1}{c|}{Transcription} & \multicolumn{1}{c|}{Text} & \multicolumn{1}{c|}{Recognition} & \multicolumn{1}{l}{Understanding}\\ \midrule

LectureVideoDB~\citeyearpar{Dutta2018} & \multicolumn{1}{c|}{M} & \multicolumn{1}{c|}{-} & \multicolumn{1}{c|}{M} & \multicolumn{1}{c|}{M} & \multicolumn{1}{c|}{-} & \multicolumn{1}{c|}{-} & \multicolumn{1}{c|}{-} &\multicolumn{1}{c|}{\ding{52}} & \ding{56} \\

SlideSpeech~\citeyearpar{Wang2023} & \multicolumn{1}{c|}{-} & \multicolumn{1}{c|}{-} & \multicolumn{1}{c|}{A} & \multicolumn{1}{c|}{-} & \multicolumn{1}{c|}{-} & \multicolumn{1}{c|}{M} & \multicolumn{1}{c|}{-} & \multicolumn{1}{c|}{\ding{52}} & \ding{56}\\

GoogleI/O~\citeyearpar{Chen2014} & \multicolumn{1}{c|}{A} & \multicolumn{1}{c|}{-} & \multicolumn{1}{c|}{A} & \multicolumn{1}{c|}{-} & \multicolumn{1}{c|}{-} & \multicolumn{1}{c|}{A} & \multicolumn{1}{c|}{-} & \multicolumn{1}{c|}{\ding{56}} & \ding{52}\\

LaRochelle~\citeyearpar{Nguyen2014} & \multicolumn{1}{c|}{A} & \multicolumn{1}{c|}{-} & \multicolumn{1}{c|}{A} & \multicolumn{1}{c|}{-} & \multicolumn{1}{c|}{-} & \multicolumn{1}{c|}{A} & \multicolumn{1}{c|}{-} & \multicolumn{1}{c|}{\ding{56}} & \ding{52} \\

LectureBank~\citeyearpar{Li2019} & \multicolumn{1}{c|}{M} & \multicolumn{1}{c|}{-} & \multicolumn{1}{c|}{A} & \multicolumn{1}{c|}{-} & \multicolumn{1}{c|}{-} & \multicolumn{1}{c|}{-} & \multicolumn{1}{c|}{-} & \multicolumn{1}{c|}{\ding{56}} & \ding{52} \\

TalkSumm~\citeyearpar{Lev2019} & \multicolumn{1}{c|}{-} & \multicolumn{1}{c|}{-} & \multicolumn{1}{c|}{-} & \multicolumn{1}{c|}{-} & \multicolumn{1}{c|}{-} & \multicolumn{1}{c|}{A} & \multicolumn{1}{c|}{A} & \multicolumn{1}{c|}{\ding{56}} & \ding{52} \\

AVIATE~\citeyearpar{Atri2021} & \multicolumn{1}{c|}{-} & \multicolumn{1}{c|}{-} & \multicolumn{1}{c|}{A} & \multicolumn{1}{c|}{-} & \multicolumn{1}{c|}{-} & \multicolumn{1}{c|}{A} & \multicolumn{1}{c|}{A} & \multicolumn{1}{c|}{\ding{56}} & \ding{52} \\

LPM~\citeyearpar{lee2022multimodal, Lee2023} & \multicolumn{1}{c|}{M} & \multicolumn{1}{c|}{M} & \multicolumn{1}{c|}{A} & \multicolumn{1}{c|}{-} & \multicolumn{1}{c|}{-} & \multicolumn{1}{c|}{A} & \multicolumn{1}{c|}{-} & \multicolumn{1}{c|}{\ding{56}} & \ding{52}\\

\bf M$^3$AV (Ours) & \multicolumn{1}{c|}{M} & \multicolumn{1}{c|}{-} & \multicolumn{1}{c|}{M} & \multicolumn{1}{c|}{M} & \multicolumn{1}{c|}{M} & \multicolumn{1}{c|}{M} & \multicolumn{1}{c|}{A} & \multicolumn{1}{c|}{\ding{52}} & \ding{52}\\
\bottomrule
\end{tabular}
}
\label{tab:related_datatype}
\end{table*}

\begin{table}[htbp]
\caption{Comparison with other academic lecture-based datasets in terms of data size and availability.}
\vspace{-0.2cm}
\resizebox{1\linewidth}{!}{
\begin{tabular}{l|cccc|c}
\toprule
 & \multicolumn{4}{c|}{\bf Size} & \multicolumn{1}{c}{\bf Available} \\ \midrule
 & \multicolumn{1}{c|}{\# Videos} & \multicolumn{1}{c|}{\# Hours} & \multicolumn{1}{c|}{\# Slides} & \multicolumn{1}{c|}{\# Papers} & \\ \midrule

 LectureVideoDB~\citeyearpar{Dutta2018} & \multicolumn{1}{c|}{24} & \multicolumn{1}{c|}{-} & \multicolumn{1}{c|}{5474} & \multicolumn{1}{c|}{-} & \ding{52}\\

SlideSpeech~\citeyearpar{Wang2023} & \multicolumn{1}{c|}{1705} & \multicolumn{1}{c|}{1080} & \multicolumn{1}{c|}{-} &  & \ding{52}\\

GoogleI/O~\citeyearpar{Chen2014} & \multicolumn{1}{c|}{209} & \multicolumn{1}{c|}{-} & \multicolumn{1}{c|}{-} & \multicolumn{1}{c|}{-} & \ding{52}\\

LaRochelle~\citeyearpar{Nguyen2014} & \multicolumn{1}{c|}{47} & \multicolumn{1}{c|}{65} & \multicolumn{1}{c|}{3250} & \multicolumn{1}{c|}{-} & \ding{56} \\

LectureBank~\citeyearpar{Li2019} & \multicolumn{1}{c|}{1352} & \multicolumn{1}{c|}{-} & \multicolumn{1}{c|}{51939} & \multicolumn{1}{c|}{-} & \ding{52}\\

TalkSumm~\citeyearpar{Lev2019} & \multicolumn{1}{c|}{1716} & \multicolumn{1}{c|}{-} & \multicolumn{1}{c|}{-} & \multicolumn{1}{c|}{1716} & \ding{52}\\

AVIATE~\citeyearpar{Atri2021} & \multicolumn{1}{c|}{8201} & \multicolumn{1}{c|}{\textasciitilde 2300} & \multicolumn{1}{c|}{-} & \multicolumn{1}{c|}{8201} & \ding{52} \\

LPM~\citeyearpar{lee2022multimodal, Lee2023} & \multicolumn{1}{c|}{334} & \multicolumn{1}{c|}{187} & \multicolumn{1}{c|}{9031} & \multicolumn{1}{c|}{-} & \ding{52}\\

\bf M$^3$AV (Ours) & \multicolumn{1}{c|}{1113} & \multicolumn{1}{c|}{367} & \multicolumn{1}{c|}{24956} & \multicolumn{1}{c|}{767} & \ding{52}\\
\bottomrule
\end{tabular}
}
\label{tab:related_datasize}
\end{table}

The datasets based on academic lectures mainly fall into two categories: measuring the model's ability to recognize multimodal content and evaluating its capacity to capture academic knowledge. In Table~\ref{tab:related_datatype}, we list the data types and dataset tasks of the relevant academic lecture-based datasets.

\paragraph{Multimodal Content Recognition.} LectureVideoDB~\cite{Dutta2018} exploits frames from lecture videos to test the model's ability to perform handwritten and scene text recognition. SlideSpeech~\cite{Wang2023} enriches the audio-visual corpus with synchronized slides to provide additional textual information. It is worth noting that M$^3$AV dataset provides both OCR data that is fully manually labelled and speech data that combines multiple ASR system outputs and manual labelling, which is completely qualified to cover tasks in the aforementioned datasets.

\paragraph{Academic Knowledge Understanding.} LectureBank~\cite{Li2019} manually annotates the prerequisite relationships between course concepts to carry out prerequisite chain learning. GoogleI/O~\cite{Chen2014} and LaRochelle~\cite{Nguyen2014} both study video-level retrieval using presentation or lecture videos. AVIATE~\cite{Atri2021} adopts multimodal information from academic presentation videos to conduct abstract summarization. Similarly, TalkSumm~\cite{Lev2019} uses a scalable annotation method to construct large-scale paper summaries from corresponding video transcription. LPM~\cite{lee2022multimodal, Lee2023} introduces crossmodal retrieval and generation tasks around its aligned slides and spoken language. Our M$^3$AV dataset contains annotated video data for both slides and speeches, which implies little knowledge loss. The addition of paper documentation information also greatly complements the knowledge information. Furthermore, a challenging Slide and Script Generation task is developed to perform the inter-transformation of two kinds of academic materials. 

As illustrated in Table~\ref{tab:related_datatype}, our dataset contains the most complete and human-annotated resources of slide, speech, and paper, thus supporting not only the recognition of multimodal content but also the comprehension of high-level academic knowledge. At the same time, the size of our dataset is also relatively rich while accessible as shown in Table~\ref{tab:related_datasize}.

\section{Summary of \texorpdfstring{M$^3$AV}{M\textasciicircum 3AV} Dataset}

\begin{table}[t]
\caption{Duration of videos from different sources in each data partition.}
\vspace{-0.2cm}
\renewcommand{\arraystretch}{0.7}
\centering
\resizebox{1\linewidth}{!}{
\begin{tabular}{l c c c c c c}
\toprule
\bf Sets & \bf CHI & \bf Ubi & \bf NIH & \bf IPP & \bf MLS & \bf Total \\
\midrule
Train & 44.1 & 7.9 & 188.2 & 34.0 & 21.1 & 295.4 \\
\midrule
Dev & 4.4 & 1.0 & 23.6 & 3.6 & 2.5 & 35.1 \\
\midrule
Test & 4.7 & 1.5 & 23.5 & 4.1 & 2.6 & 36.4 \\
\bottomrule
\end{tabular}
}
\label{tab:sets}
\vspace{-0.4cm}
\end{table}

\begin{figure*}[htbp]
    \centering
    \includegraphics[width=0.9\linewidth]{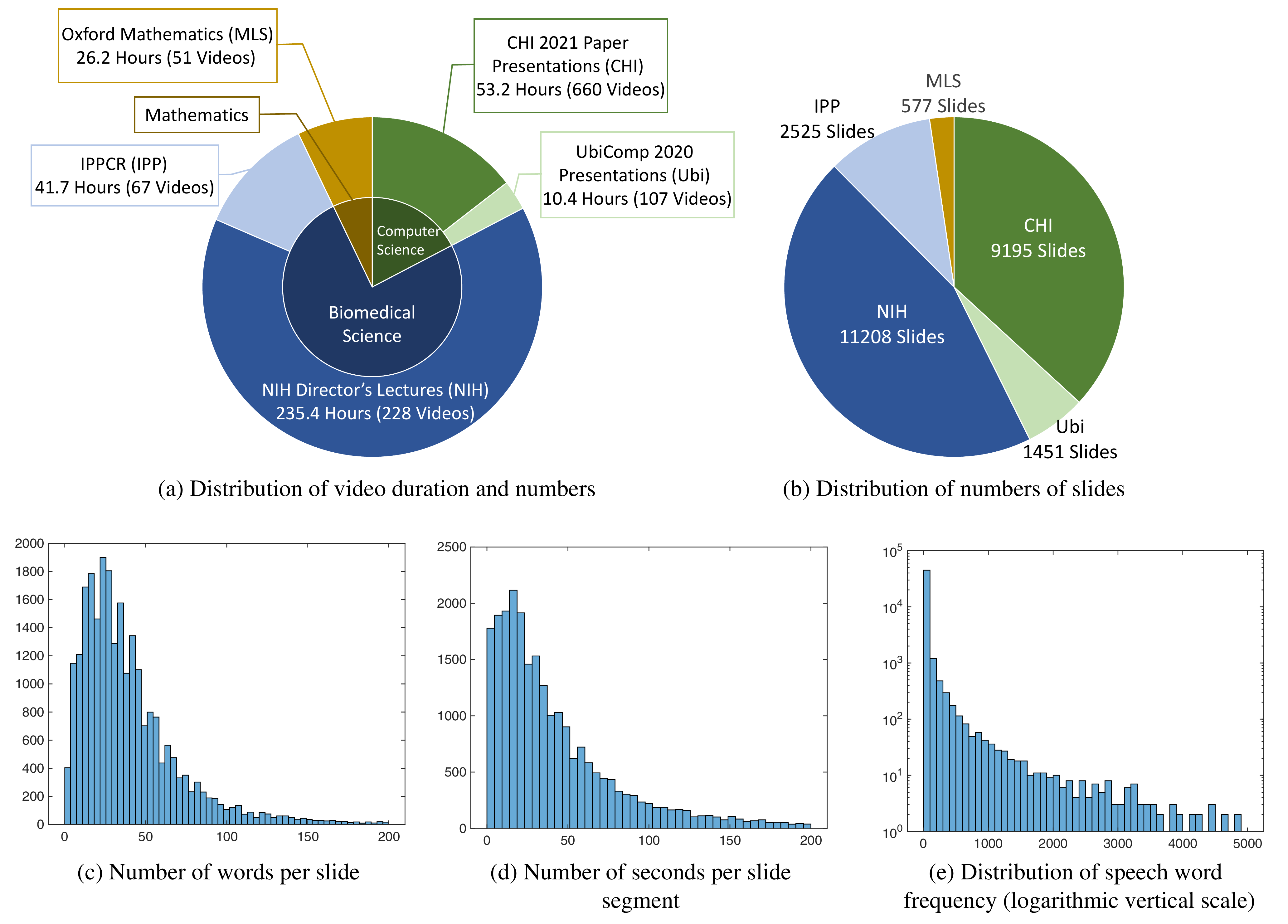}
    \vspace{-0.2cm}
    \caption{Statistics of our dataset. (a) shows video duration and numbers. (b) shows the number of slides. (c) shows the number of words per slide. (d) shows the duration per slide segment. (e) shows the speech word frequency.}
    \label{fig:statistics}
    \vspace{-0.3cm}
\end{figure*}

\subsection{Metadata}
We treat each video as a unit and rename each video in the form of "\{source\}-\{id\}". For the raw data, we provide the YouTube URL of each video for users to download.
\paragraph{Speech Annotation.} The speech annotations are saved in JSON files. Each unit contains the start and end timestamps, the transcribed text in the spoken and written form, and the corresponding word-level timestamps.
\paragraph{OCR Annotation.} In OCR annotations, the coordinates and manually corrected text (formulae are in LaTeX format) of each block are provided in JSON files. Results of merging (detailed in Appendix~\ref{app:seg_rules}) are also provided.
\paragraph{Dataset Split.} The division of the sets is determined by the speech data, as the duration can measure the amount of information consistently. The video durations are illustrated in Table~\ref{tab:sets}.

\subsection{Characteristics of Speech Data}\label{sec:speech_analysis}

\paragraph{Variety of Academic Fields.} As shown in Figure~\ref{fig:statistics}\nolinebreak(a), M$^3$AV contains academic presentations and talk videos from three different fields including computer science, biomedical science, and mathematics. The Conference on Human Factors in Computing Systems (CHI) 2021 Presentations and ACM International Conference on Ubiquitous Computing (Ubi) 2020 Presentations focus on research in computer science. The National Institutes of Health Director's Wednesday Afternoon Lectures (NIH) presents weekly science talks given by top biomedical scientists. The Introduction to the Principles and Practice of Clinical Research (IPP) teaches how to conduct clinical research effectively and safely. Oxford Mathematics (MLS) contains mathematics lectures. The total number of videos is 1113, with 366.9 hours of speech.
\paragraph{Abundance of Rare Words.} The size of our spoken form word table is 47865, while the word frequency within 1k reaches 47483 words (99.20\%) as shown in Figure~\ref{fig:statistics}\nolinebreak(e). 
It effectively represents the typical scenario met in academic presentations where new terminologies constantly appear and are crucial to the overall understanding. 

\paragraph{Quality of Transcription.} M$^3$AV is labelled by combining multiple high-performing ASR systems together with manual labels. As human often fails to label rare terminologies, the assistance of a high-performing ASR is indispensable. The multimodal labelling pipeline is described in Section~\ref{sec:speechpipeline}.

\subsection{Characteristics of Visual Data}
\paragraph{Suffient OCR Text Labelling.} The total number of slides has reached 24956 as illustrated in Figure~\ref{fig:statistics}\nolinebreak(b). The average number of words per slide is 40.96 as shown in Figure~\ref{fig:statistics}\nolinebreak(c). We can observe that there is a sufficient number of slides and the density of the words is high, which indicates the richness of the textual information in our slides. The duration of every slide segment is 50.67 as indicated in Figure~\ref{fig:statistics}\nolinebreak(d). 

\paragraph{Complex Handwriting and Formulae.} Significant numbers of pages and blocks in Table~\ref{tab:complex} show the enrichment of our complex texts. Furthermore, handwriting and formula texts which both contain lots of presentation content and academic knowledge have already been processed into standard text or LaTeX format and thus can be directly utilised by large language models.

\begin{table}[t]
\renewcommand{\arraystretch}{0.7}

\caption{Statistics of handwriting and formulae. Blocks that appear on the same page as the complex block are also counted into "\# Blocks" due to the same processing.}
\vspace{-0.2cm}
\centering
\begin{tabular}{c c c}
\toprule
\bf Sources & \bf \# Pages & \bf \# Blocks* \\
\midrule
CHI & 58 & 656 \\
\midrule
Ubi & 63 & 906 \\
\midrule
NIH & 53 & 695 \\
\midrule
IPP & 21 & 224 \\
\midrule
MLS & 577 & 5736 \\
\midrule
Total & 772 & 8217 \\
\bottomrule
\end{tabular}
\label{tab:complex}
\end{table}

\section{Data Creation Pipeline}

\subsection{Data Collection}

Videos of open-source conferences and lectures are collected from YouTube, together with descriptions and subtitles.
Papers for computer science videos are downloaded and subsequently parsed using science-parse\footnote{\url{https://github.com/allenai/science-parse}}.

\subsection{Speech Transcription} 
\label{sec:speechpipeline}
\begin{figure}[htbp]
    \centering
    \includegraphics[width=0.9\linewidth]{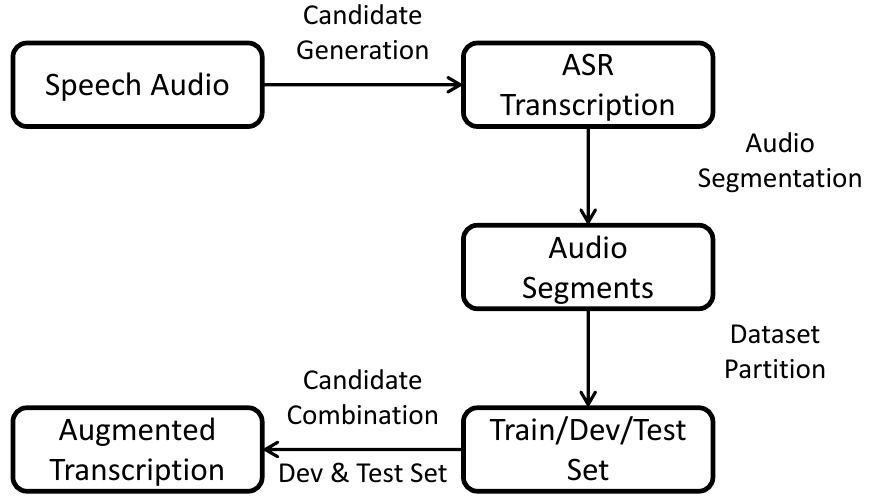}
    \caption{Diagram illustration of the process of creating speech transcription.}
    \label{fig:speech_ann}
    \vspace{-0.3cm}
\end{figure}

The diagram of creating speech transcription is shown in Figure~\ref{fig:speech_ann}, comprising four steps.

\subsubsection{Step 1: Candidate Generation}\label{sec:candidate_generation}

The downloaded subtitles are not high-quality as expected by our observation. Therefore, we introduce expert ASR systems following~\citet{Zhang2022}. In particular, we select Microsoft STT\footnote{\url{https://azure.microsoft.com/en-us/products/ai-services/speech-to-text}}, and Whisper-large-v2~\cite{Radford2023}\footnote{\url{https://github.com/openai/whisper}}. They both achieve a WER of less than 10\% or even 5\% on multiple test sets indicated in the SpeechColab ASR Benchmark (EN)\footnote{\url{https://github.com/SpeechColab/Leaderboard}} in Oct. 2022.

The Microsoft transcriptions in written and spoken form\footnote{``Spoken form'' refers to the representation of language as it is actually spoken or pronounced, while ``written form'' refers to the representation of language in its written or textual format.} are denoted as $M_w$ and $M_s$ respectively, and the Whisper transcriptions result in written form are denoted as $W_w$ (only the written form exists in the Whisper result). 

\begin{figure}[htbp]
    \centering
    \includegraphics[width=0.8\linewidth]{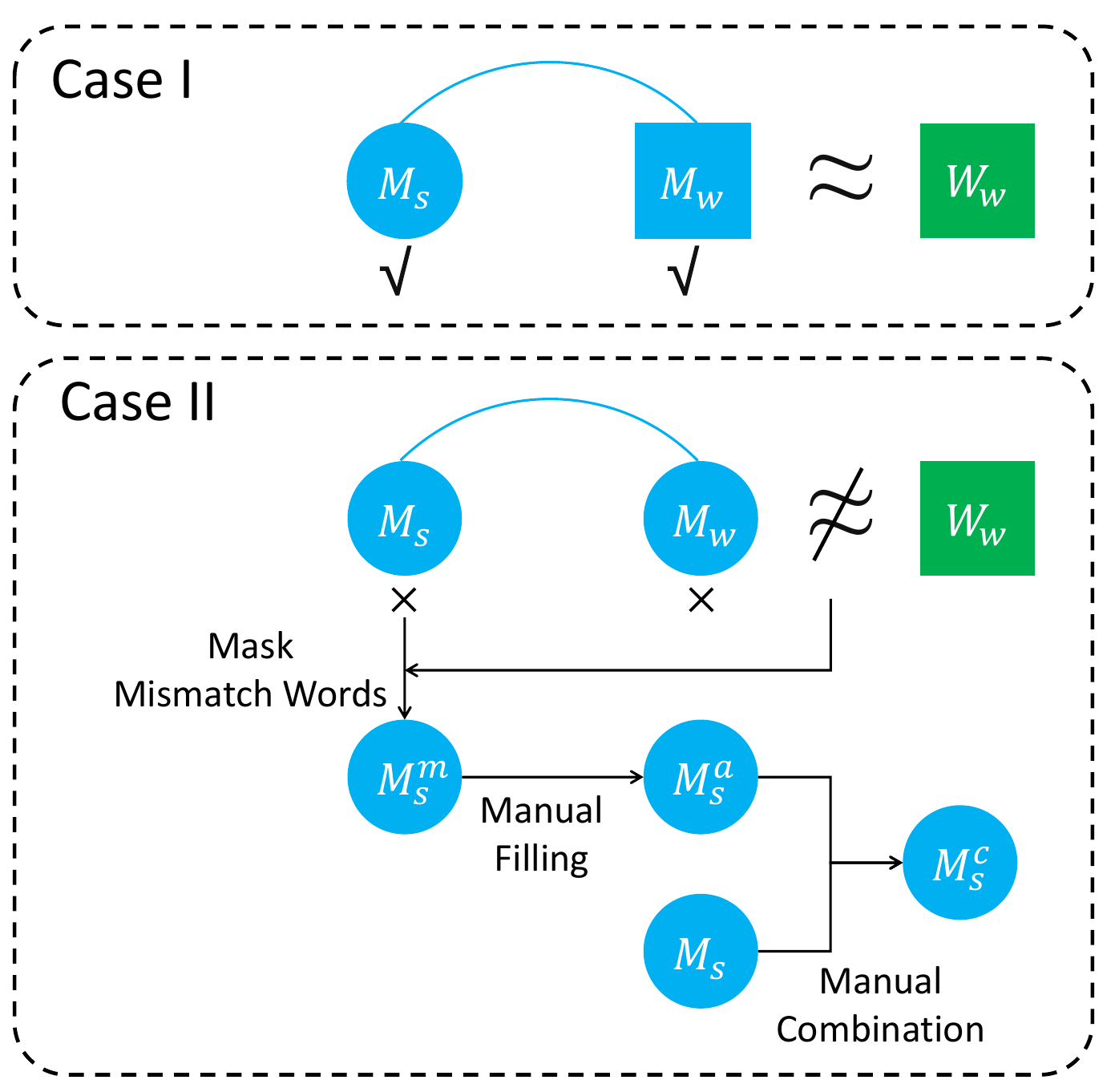}
    \caption{Diagram illustration of the process of candidate combination.}
    \label{fig:combination}
    \vspace{-0.3cm}
\end{figure}

\subsubsection{Step 2: Audio Segmentation}

Most ASR models expect that the input audio comes in the form of a relatively short duration, usually up to a few seconds in length. Therefore, we conduct re-segmentation using the word-level timestamps in $M_w$. The split is allowed to be in silence and punctuation. And the duration of each segment is kept to 10 seconds or less. Detailed rules can be found in Appendix~\ref{app:seg_rules}.

\begin{figure*}[t]
    \centering
    \includegraphics[width=0.9\textwidth]{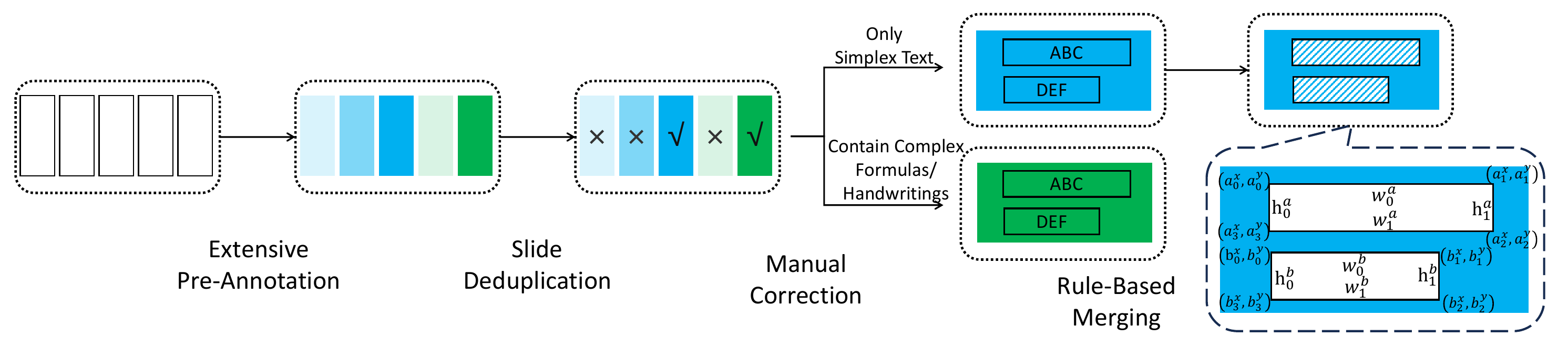}
    \caption{Diagram illustration of the process of slide annotation. Shades of the same colour represent the amount of slide content in the same segment. For example, the dark green page adds some content to the light green page. The right sign (\ding{52}) represents reservation, while the wrong sign (\ding{53}) represents discarding.}
    \label{fig:slide_ann}
    \vspace{-0.3cm}
\end{figure*}

\subsubsection{Step 3: Dataset Partition}
We divide all the videos into training, development, and test sets, with a ratio of roughly 8:1:1 for total duration. We also avoid a speaker appearing in different sets, which is identified by the introductory information and description text.

\subsubsection{Step 4: Candidate Combination}

For the training set, we directly adopt the Microsoft transcription $M_w$ and $M_s$, because it shows high quality verified by checking against manual labelling on a subset and has both spoken and written form. For the development and test set, we design the candidate combination process shown in Figure~\ref{fig:combination}. 

In Case I where $M_w$ is "approximately equal"\footnote{It forgives misalignments due to pauses and common words, as they are mainly short-pronounced, and $M_w$ and $M_s$ have high confidence in them.} to $W_w$, $M_s$ and $M_w$ will be directly adopted.

In Case II where $M_w$ is not "approximately equal" to $W_w$, we design the following steps:
\begin{compactenum}
    \item We use the misalignment to mask the words in $M_s$ as \texttt{[???]}.
    \item Annotators are assigned to fill in the masks. The result, $M_s^{a}$ can be regarded as a type of data augmentation.
    \item Annotators are instructed to combine $M_s$ and $M_s^{a}$ to provide the final corrected results $M_s^{c}$. 
\end{compactenum}
The postprocessing of $M_s^c$ is detailed in Appendix~\ref{app:post_comb}. The manual labelling process is detailed in Appendix~\ref{app:man}. The comparison with manual labelling on a subset for the training set and the manual processing for the dev/test set ensure the accuracy and consistency of the speech annotation.

\subsection{Slide Annotation}

The diagram of creating slide annotation is shown in Figure~\ref{fig:slide_ann}, comprising four steps.

\subsubsection{Step 1: Extensive Pre-Annotation}
We conduct frame sampling for the video at 1 frame per second and use PaddleOCR\footnote{\url{https://github.com/PaddlePaddle/PaddleOCR}} for large-scale text detection and recognition. PaddleOCR is selected for its completeness and high accuracy of the recognition results with little sticking between words.

\subsubsection{Step 2: Slide Deduplication}
Since the frames are repetitive, we first perform automatic deduplication. Specifically, we remove uninformative frames and blocks and filter out frames with too few or too many blocks. Next, we compare the OCR contents of neighbouring frames and only keep the latter one when the content of the former is entirely included. Detailed criteria are shown in Appendix~\ref{app:dedup}. After completing the automatic deduplication, manual deduplication is performed to retain the last frame in each segment.

\subsubsection{Step 3: Manual Correction}

For frames with simple text, corrections are directly executed by experienced researchers for the location and content of the OCR results. Note that blocks of presentation-irrelevant text such as URLs and email addresses are omitted. 
For frames with complex handwriting or formulae, we pre-label them using the Mathpix OCR API\footnote{\url{https://mathpix.com/ocr}}. Then, the annotator performs careful corrections for the OCR result due to their complexity. The formulae are labelled in the format of in-line LaTeX. The details of manual labelling are shown in Appendix~\ref{app:man}. To control the quality of labelling, we sampled the results of 9 annotators individually and checked whether 120 of the annotated images had a high number of missing or incorrect items, with no more than 10 wrong items.

\subsubsection{Step 4: Rule-Based Merging}

To provide comprehensive semantic information, we devise three rules to perform merging\footnote{As demonstrated in the top part of Figure~\ref{fig:dataset}, the recognition results of the general OCR model are in rows. It is required to merge these lines into paragraphs to provide complete semantic information.}:
\begin{compactitem}
\item Similar Height:
\vspace{-0.2cm}
    \begin{equation}
    \begin{aligned}
        \max(h_0^a, h_0^b) &- \min(h_0^a, h_0^b) \\ & \leq a \times \max(h_0^a, h_0^b)
    \end{aligned}
    \end{equation}
\vspace{-0.2cm}
\item High Overlap in the Horizontal Direction:
\vspace{-0.2cm}
    \begin{equation}
    \begin{aligned}
        \min(& a_2^x, b_1^x) - \max(a_3^x, b_0^x) \\ & \geq b \times \min(b_1^x - b_0^x, a_2^x - a_3^x)
    \end{aligned}
    \end{equation}
\vspace{-0.2cm}
\item Proximity of the Upper and Lower:
\vspace{-0.2cm}
    \begin{equation}
    \begin{aligned}
        b_0^y - a_3^y \leq c \times \min(h_0^a, h_0^b)
    \end{aligned}
    \end{equation}
\end{compactitem}
\vspace{-0.2cm}
where the notations are shown as Figure~\ref{fig:slide_ann}. We set $a=0.8$, $b=0.8$, and $c=0.6$.

\section{Benchmarks and Experiments}

\begin{table*}
\caption{\label{Tab: ASR result} Evaluation results on ASR and CASR tasks. }
\vspace{-0.2cm}
\centering
\resizebox{1\linewidth}{!}{
\begin{tabular}{l l c c c c c c}
\toprule
\multirow{2} {*}{\bf Model} & \multirow{2} {*}{\bf Biasing List}& \multicolumn{3}{c}{\bf Dev} & \multicolumn{3}{c}{\bf Test}\\  \cmidrule{3-5} 
 \cmidrule{6-8}
 & & CER ($\downarrow$) & WER ($\downarrow$) & BWER ($\downarrow$) & CER ($\downarrow$) & WER ($\downarrow$)& BWER ($\downarrow$)\\
\midrule
\multicolumn{8}{c} {\emph{ASR}}\\
\midrule
AED & - & 4.6 & \bf 10.0 & \bf 35.5 & 4.5 & \bf 10.0 & \bf 35.6 \\
RNN-T & - & 4.7 & 10.3 & 36.8 & 4.7 & 10.6 & 38.0 \\
\midrule
\emph{Whisper-v3$_{\textsc{1.5B}}$} & \emph{-} & \emph{3.8} & \emph{7.1} & \emph{19.4} & \emph{3.8} & \emph{7.2} & \emph{18.9} \\
Wav2Vec2+CTC$_{\textsc{0.3B}}$ & - & 10.6 & 28.3 & 60.1 & 10.6 & 28.5 & 62.4 \\
Conformer+Trans.$_{\textsc{0.6B}}$ & - & 4.6 & 10.7 & 37.8 & 4.6 & 10.5 & 38.2 \\
Conformer+CTC$_{\textsc{0.6B}}$ & - &  \bf 4.4 & 11.1 & 40.6 & \bf 4.3 & 11.1 & 42.2 \\
SeamlessM4T-v2$_{\textsc{2.3B}}$  & - & 12.4 & 17.1 & 36.7 & 12.4 & 18.6 & 40.7 \\
\midrule
\multicolumn{8}{c} {\emph{CASR}}\\
\midrule
RNN-T + TCPGen & Random Distractors & 4.7 & 9.7 & 26.7 & 4.7 & 9.7 & 27.1 \\
RNN-T + TCPGen & OCR + Same Topic Rare Words& 4.8 & 9.8 & 30.5 & 4.7 & 10.0 & 31.7 \\
RNN-T + TCPGen & OCR & \bf 4.5 & \bf 9.4 & \bf 22.9 & \bf 4.6 & \bf 9.6 & \bf 25.0\\
\bottomrule
\end{tabular}
}
\vspace{-0.2cm}
\end{table*}

\subsection{ASR and CASR Task}
\subsubsection{Task Description}

E2E approaches have gained substantial attention for ASR, which enable direct mapping from raw speech audio to its transcription. However, they often struggle with the accurate recognition of rare words that carry important information for understanding. CASR aims to improve the model's recognition performance for these rare words by incorporating them into the biasing list.

\subsubsection{Benchmark Systems}

We adopt two widely-used benchmark systems for ASR and CASR, namely Attention-based Encoder-Decoder (AED)~\cite{Chan2016} and RNN-Transducer (RNN-T)~\cite{Graves2012}. Both systems employ a Conformer~\cite{Gulati2020} encoder. The AED system is jointly trained and decoded with Connectionist Temporal Classification (CTC)~\cite{Graves2006, Watanabe2017}.

For ASR, we also introduce a variety of state-of-the-art pretrained ASR models. Note that the results of Whisper-v3 are for reference only, as the Whisper-series model is involved in dataset construction. The performance of the other open-source ASR models can be fairly evaluated, given that our videos are properly transcribed for the first time, avoiding the issue of training data contamination.  These pretrained models are selected from the SpeechColab ASR Benchmark (EN)\footnote{\url{https://github.com/SpeechColab/Leaderboard}} and Hugging Face Community\footnote{\url{https://huggingface.co/models?pipeline_tag=automatic-speech-recognition&sort=trending}} which include Whisper-v3~\cite{Radford2023}, Wav2Vec2+CTC~\cite{grosman2021xlsr53-large-english}, Conformer+Transducer\footnote{\url{https://catalog.ngc.nvidia.com/orgs/nvidia/teams/nemo/models/stt_en_conformer_transducer_xlarge}}, Conformer+CTC\footnote{\url{https://catalog.ngc.nvidia.com/orgs/nvidia/teams/nemo/models/stt_en_conformer_ctc_xlarge}} and SeamlessM4T-v2~\cite{barrault2023seamless}\footnote{\url{https://huggingface.co/facebook/seamless-m4t-v2-large}}.

For CASR, the Tree-Constrained Pointer Generator (TCPGen)~\cite{Sun2023} with GNN tree encodings~\cite{Sun2023a} is used which employs a combination of symbolic prefix-tree search and a neural pointer generator for contextual biasing. More details are illustrated in Appendix~\ref{app:asr} and Appendix~\ref{app:casr}.

\subsubsection{Experimental Results}

In addition to the character error rate (CER) and word error rate (WER) for ASR, we use biasing WER (BWER) for CASR. BWER is defined as the total rare word errors divided by the total number of rare words, including insertions of rare words. Table~\ref{Tab: ASR result} shows the experimental results. The AED and RNN-T systems achieve around 10.0\% WER on both the Dev and Test sets. However, such end-to-end models suffer from rare word recognition as reflected by the BWER where a more than two times increase in the error rate is observed comparing BWER to the WER. 

When 1000 distractors\footnote{We select a certain number of significant words, which are the words in the bias list we aim to recognize accurately. When decoding, a particular utterance only includes a few of these words, while the rest will appear as distractors. The use of distractors follows \citeauthor{Le2021,Sun2022}.} are randomly selected from all rare words and are added to the biasing list, significant improvements in rare word recognition can be observed. When using OCR information together with rare words from the same topic to form the biasing list of each utterance during decoding, the recognition performance slightly deteriorates. This is because strongly related words are included in the biasing list, which may cause more confusion since words with related meanings often have similar pronunciations and word forms. When only OCR words are used, we achieve the best performance of the CASR task. In conclusion, by using TCPGen utilizing the OCR information, we achieve a relative BWER decrease of 37.8\% and 34.2\% on dev and test sets respectively.

\subsection{Spontaneous TTS Task}
\subsubsection{Task Description}

Spontaneous Text-to-Speech (TTS)~\cite{Guo2021,Yan2021} focuses on generating speech with a natural and conversational session. Most TTS systems use clean speech. This form makes it easier to extract pronunciation features, while it places higher requirements on the quality of the training audio and often leads to a discernible gap between the synthesized output and authentic speech. We are devoted to using real speech data in M$^3$AV for the TTS system to produce speech that aligns more closely with natural conversational patterns, as it is spontaneous and has regulated pronunciations.

\subsubsection{Benchmark Systems}

Multi-codebook vector quantized TTS (MQTTS) \cite{Chen2023} is introduced as our spontaneous TTS baseline. It employs a two-stage training architecture. In the first stage, a quantizer is trained to discretize the raw waveform into a sequence of codes, minimizing quantization and reconstruction losses. A discriminator guides reconstruction via adversarial learning. In the second stage, the fixed quantizer is combined with a Transformer that performs autoregressive generation on the code sequence, conditioned on a speaker embedding for multi-speaker capability. More details can be found in Appendix~\ref{app:tts}.

\subsubsection{Experimental Results}

\begin{table}
\renewcommand{\arraystretch}{0.9}
\caption{\label{Tab: TTS result} Evaluation results on Spontaneous TTS task. ``GT'' denotes the ground truth.}
\vspace{-0.3cm}
\centering
\resizebox{1\linewidth}{!}{
\begin{tabular}{l c c c c c c c}
\toprule 
\multirow{2} {*}{\bf Model} & \multicolumn{3}{c}{\bf Objective} & \multicolumn{1}{c}{\bf Subjective}\\  \cmidrule{3-5} 
 \cmidrule{6-8}
 & & FFE ($\downarrow$) & MCD ($\downarrow$) & MOS ($\uparrow$)\\
\midrule
\emph{GT} & & \emph{-} & \emph{-} & \emph{3.86 $\pm$ 1.02}\\
Bark & & 0.4269 & 10.57 & 3.31 $\pm$ 1.02\\
SpeechT5 & & 0.4153 & 9.43 & 3.43 $\pm$ 1.09\\
MQTTS & & \textbf{0.3518} & \textbf{9.40} & \textbf{3.67 $\pm$ 0.94}\\
\bottomrule
\end{tabular}
}
\vspace{-0.4cm}
\end{table}

For the objective metrics, we adopt the F0 Frame Error (FFE)~\cite{Chu2009} and Mel-Cepstral Distortion (MCD)~\cite{Kubichek1993}. 
For the subjective metrics, the Mean Opinion Score (MOS) test on a scale of 1 to 5 is performed, which is detailed in Appendix~\ref{app:man}. We randomly choose 20 samples and generate speech segments with the MQTTS model and well-known pre-trained models, including Bark\footnote{\url{https://huggingface.co/suno/bark}} and SpeechT5~\cite{Ao2022}\footnote{\url{https://huggingface.co/microsoft/speecht5\_tts}}. Table~\ref{Tab: TTS result} shows the experimental results, in which the MQTTS model shows the best performance within all the evaluation metrics. It indicates that the real speech in our dataset can drive AI systems to simulate more natural speech.

\subsection{Slide and Script Generation Task}
\subsubsection{Task Description}

\begin{table*}[t]
\renewcommand{\arraystretch}{0.95}
\caption{Evaluation results on SSG tasks. The upper part of "Slide$\rightarrow$Script" shows cascading pipelines, while the lower part shows integrated systems.}
\vspace{-0.2cm}
\centering
\resizebox{1\linewidth}{!}{
\begin{tabular}{l c c c c c c}
\toprule
\bf Model & \bf ROUGE-1 ($\uparrow$) & \bf ROUGE-2 ($\uparrow$) & \bf ROUGE-L ($\uparrow$) & 
 \bf BERTScore ($\uparrow$) & \bf BARTScore ($\uparrow$) & \bf Human ($\uparrow$)\\
\midrule 
\multicolumn{7}{c} {\emph{Slide $\longrightarrow$ Script}}\\
\midrule
LLaMA-2$_{\textsc{7B}}$+OCR & 24.6 & 5.6 & 21.6 & 0.073 & -4.970 & 3.6 \\
LLaMA-2$_{\textsc{13B}}$+OCR & \bf 26.9 & \bf 6.4 & 23.6 & \bf 0.102 & -4.823 & 3.9 \\
GPT-4+OCR & 26.2 & 5.8 & \bf 23.7 & 0.085 & \bf -4.621 & 4.6 \\
\midrule
InstructBLIP$_{\textsc{7B}}$ & 24.3 & 2.2 & 17.9 & 0.041 & -5.504 & 2.4 \\
InstructBLIP$_{\textsc{13B}}$ & 24.7 & 2.3 & 18.2 & 0.046 & -5.514 & 2.5 \\
GPT-4V & 26.4 & 6.2 & \bf 23.7 & 0.092 & -4.630 & \bf 4.7 \\
\midrule
\multicolumn{7}{c} {\emph{Script $\longrightarrow$ Slide}}\\
\midrule
LLaMA-2$_{\textsc{7B}}$ & 20.3 & 6.2 & 17.3 & 0.064 & -6.689 & 3.3 \\
LLaMA-2$_{\textsc{13B}}$ & 23.3 & 7.5 & 19.8 & \bf 0.092 & -6.484 & 3.8\\
GPT-4 & \bf 26.8 & \bf 7.7 & \bf 23.0 & 0.077 & \bf -6.136 & \bf 4.8 \\
\bottomrule
\end{tabular}
}
\label{tab:rouges_models}
\end{table*}

\begin{table*}[t]
\renewcommand{\arraystretch}{0.95}
\caption{Performance improvements of LLaMA-2$_{\textsc{7B}}$ brought by retrieving paper information. ``Subset'' denotes that only Computer Science videos are contained in all sets for they are the only ones with downloadable papers.}
\vspace{-0.2cm}
\centering
\resizebox{1\linewidth}{!}{
\begin{tabular}{l c c c c c c}
\toprule
\bf Range & \bf ROUGE-1 ($\uparrow$) & \bf ROUGE-2 ($\uparrow$) & \bf ROUGE-L ($\uparrow$) & 
 \bf BERTScore ($\uparrow$) & \bf BARTScore ($\uparrow$) & \bf Human ($\uparrow$)\\
\midrule 
\multicolumn{7}{c} {\emph{Slide $\longrightarrow$ Script}}\\
\midrule
Subset & 24.9 & 6.4 & 21.8 & 0.114 & -5.110 & 3.7 \\
~~ + Paper & \bf 29.2 & \bf 8.7 & \bf 25.8 & \bf 0.161 & \bf -4.806 & \bf 4.1 \\
\midrule
\multicolumn{7}{c} {\emph{Script $\longrightarrow$ Slide}}\\
\midrule
Subset & 20.9 & 6.8 & 18.1 & \bf 0.061 & -6.737 & 3.3 \\
~~ + Paper & \bf 22.3 & \bf 7.9 & \bf 19.3 & \bf 0.061 & \bf -6.651 & \bf 3.6 \\
\bottomrule
\end{tabular}
}
\label{tab:rouges_llama}
\vspace{-0.4cm}
\end{table*}

We define the task formally as follows. Let $I$, $T_{i}$ and $T_{s}$ be the slide picture, slide text and the speech transcription of every segment, respectively. For the ``Slide$\rightarrow$Script'' task, we need to generate $T_{s}$ based on $I$. Note that $T_{i}$ can be viewed as the intermediate result generated from $I$ by the OCR model. For the ``Script$\rightarrow$Slide'' task, we need to generate $T_{i}$ based on $T_{s}$.
The Slide and Script Generation (SSG) Task is designed to promote AI models to understand and reconstruct advanced academic knowledge, thus assisting researchers in creating slides and presentations. The selected paper sentences are provided for retrieval-augmented generation (RAG), supplementing rich knowledge. Such models with academic comprehension can facilitate researchers to handle frequently updated academic materials to acquire knowledge and innovation efficiently.

\subsubsection{Benchmark Systems}

Having witnessed the excellent performance of large language models (LLMs) and large multimodal models (LMMs) nowadays, we explore their capabilities there. For the "Slide$\rightarrow$Script" task, we adopt multimodal cascading pipelines and integrated systems. The cascading pipeline consists of an OCR model to extract the text (we directly adopt the OCR annotations) and a language model to generate the answer. Specifically, the cascading pipelines include fine-tuned LLaMA-2~\cite{Touvron2023} and GPT-4~\cite{OpenAI2023} using slide OCR text and the integrated systems include fine-tuned InstructBLIP~\cite{Dai2023} and GPT-4V~\cite{OpenAI2023a, OpenAI2023b}. For the "Script$\rightarrow$Slide" task, we adopt fine-tuned LLaMA-2 and GPT-4. For the selection of related paper sentences, We calculate the similarity between the script text and each paper sentence and filter from them. More details can be found in Appendix~\ref{app:slide}.

\subsubsection{Experimental Results}

ROUGE~\cite{Lin2004}, BERTScore~\cite{zhang2019bertscore}, and BARTScore~\cite{yuan2021bartscore} results are reported in Table~\ref{tab:rouges_models} and Table~\ref{tab:rouges_llama}. We also perform manual scoring, where each sample is scored on a scale from 1 (worst) to 5 (best) by 3 annotators on multiple dimensions, including coherence, consistency, fluency and relevance~\cite{fabbri2021summeval}. We show their average values, while the detailed values are in Appendix~\ref{ssg_detail_score}. Scoring is performed on a subset of size 100, detailed in Appendix~\ref{app:man}. 

We get significant observations: 1)~The open-source models (LLaMA-2, InstructBLIP) show a limited performance improvement when raised from 7B to 13B. Their performances are far from the closed-source models (GPT-4 and GPT-4V). We believe that high-quality pre-training data, e.g., informative corpus and visual QA data which encapsulate multimodal information, is required to enhance their SSG performances beyond just boosting the model size. 2)~The latest LMM (GPT-4V) has already exceeded the cascaded pipeline composed of unimodal expert models. It suggests that the LMM not only maintains the ability to process textual information but also possesses multi-sensory capabilities, such as the perception and recognition of the slides~\cite{Yang2023}. 3)~RAG substantially enhances the generation, as shown in Table~\ref{tab:rouges_llama}. Qualitative analyses in Appendix~\ref{sec:cases} provide a more fine-grained view.

As for future development, we are urged to strengthen the open-source model's ability to understand high-level knowledge and perceive multimodal information. They can be achieved by increasing the model size and supplying high-quality data. The issue that the open-source multimodal models struggle with recognising text in images can be mitigated by increasing the OCR input. Furthermore, how to effectively bring in external knowledge is an essential topic.
\section{Conclusion}
We release the Multimodal, Multigenre, and Multipurpose Audio-Visual Dataset with Academic Lectures (M$^3$AV) covering a range of academic fields. This dataset contains manually annotated speech transcriptions, slide text, and additional extracted papers, providing a basis for evaluating AI models for recognizing multimodal content and understanding academic knowledge. We detail the creation pipeline and conduct various analyses of the dataset. Furthermore, we build benchmarks and conduct experiments around the dataset. We find there is still large room for existing models to improve perceptions and understanding of academic lecture videos.
\section*{Limitations}

There are some limitations in our work. First, there may exist biases in the dataset, such as the types of presentations and the demographics of the speakers. It is hard to completely avoid biases, although we have incorporated a diverse range of data sources. The users of the dataset should be aware of the impact brought to the AI models by these biases.

Secondly, the data we collected has limitations in terms of academic domains and multilingualism. We will consider adding videos from more research directions, such as humanities disciplines (e.g. economics, law, and sociology) to an extended dataset in future work. And we also plan to collect more data in other languages to prompt multilingual academic research. Nevertheless, we believe our current collection would already be useful for the community to develop some initial approaches.

Thirdly, we have some deficiencies in mining and labelling for visual information of the dataset. The illustrations in the video frames are not extracted separately like \citet{lee2022multimodal, Lee2023}. And, the talking heads of the speakers are not picked up, which can be utilised to research the generation of virtual presenters. In summary, further exploitation of the visual elements in the slides would be a future extension of the dataset.

Finally, the benchmark systems for the Slide Generation task do not involve the generation of visual pictures, instead, we only get the generated slides in the form of text. The generation of slide images requires complex multimodal models, and the corresponding evaluation metrics require more sophisticated designs.

\section*{Ethical Consideration}

This work presents a dataset based on academic speeches with manual annotation in terms of slides and speech. The presentations and talks included in the dataset, are open accessed on the YouTube website. The copyright remains with the original owners of the videos. Our dataset is provided under a Creative Commons Attribution-NonCommercial-ShareAlike 4.0 International License. The annual annotation is done by the team's researchers and workers on the Amazon Mechanical Turk. All annotators received clear annotation rules before the annotation and fair payments upon completion. No personal information is requested from any of the annotators.

\section*{Acknowledgements}
This work is supported by National Key R\&D Program of China (No.~2022ZD0162101), National Natural Science Foundation of China (No.~62106140) and STCSM (No.~21511101100, No.~22DZ2229005).

\bibliography{main}
\appendix
\newpage
\section{Creation Pipeline Details}

\subsection{Audio Segmentation Rules}~\label{app:seg_rules}

We establish the following rules of segmentation:
\begin{compactenum}
    \item Split is allowed when the silence exceeds 0.2 seconds or there is sentence break punctuation (".", "!", "?"). 
    \item If the accumulated "small segments" obtained are greater than or equal to 8 seconds or the accumulation is greater than or equal to 10 seconds after the next "small segment" is added, then the accumulated segments will form a new segment.
    \item If the silence exceeds 5 seconds, then this silence is discarded and a new segment is formed from the accumulated "small segments".
\end{compactenum}
 The first rule aims to preserve boundary words as well as the semantics of sentences inspired by LibriSpeech~\cite{Panayotov2015} and GigaSpeech~\cite{Chen2021}. The second rule controls the duration of each segment, while the third rule avoids too long silence within a segment.
 
\subsection{Auto Slide Deduplication Details}\label{app:dedup}
\begin{compactenum}
\item
For the frames that contain some OCR block representing the video cover $B_{\text{cover}}$, we delete these images directly. And we remove fixed appearing logo blocks $B_{\text{excluded}}$, such as "NIH".

\item
Then, we set a threshold $\min_{\text{alpha}}^{\text{B}}$ to filter blocks with too few letters. For the filtering process for images, we set the thresholds $\min_{\text{B}}^{\text{I}}$ and $\max_{\text{B}}^{\text{I}}$ to filter images with too few blocks and too many blocks, respectively.

\item
After the above filtering, we merge all the blocks of each image into long text by their coordinates. 

\item 
Initially, we assign the first image to segment $0$. Next, we will iterate through each image chronologically: if the error rates (ER) between the current image  (segment $i$) content and all past image contents are all greater than the threshold $\min_{\text{ER}}$, then it is considered as a new image and assigned to segment $i+1$, otherwise, it is not and kept as segment $i$.

\item 
Eventually, the last image in each segment will be the result of the deduplication since they are not repeating each other and contain the most text of the images in their segment.
\end{compactenum}

We would like to emphasize that the thresholds mentioned above depend on the characteristics of each source of videos, such as the clarity of the picture, the density of the text, and the presence of handwriting and formulae. Moreover, we adopt a modified Word Error Rate (WER) or Char Error Rate (CER) in the calculation of ER mainly considering the incremental content of slides. They are formulated as follows:
\begin{align}
    \text{WER}^{\text{mod}} &= \frac{S + D + 0 \times I}{H + S + D} \\
    \text{CER}^{\text{mod}} &= \frac{S + D + 0.1 \times I}{H + S + D}
\end{align}
where $S$, $D$, $I$, and $H$ represent the number of substitutions, deletions, insertions, and hits, respectively. The choice of WER or CER also depends on the aforementioned characteristics of the videos.

\subsection{Postprocessing for Manual Combination Results}\label{app:post_comb}

To preserve the consistency of the data after obtaining $M_s^c$ shown in Figure~\ref{fig:combination}, we perform inverse text normalization\footnote{\url{https://docs.nvidia.com/deeplearning/nemo/user-guide/docs/en/main/nlp/text_normalization/wfst/wfst_text_normalization.html\#inverse-text-normalization}}, add punctuations, and capitalize words using tools\footnote{\url{https://docs.nvidia.com/deeplearning/nemo/user-guide/docs/en/main/nlp/punctuation_and_capitalization.html}}. Finally, we obtain the word-level timestamps of $M_s^{c}$ using Montreal Forced Aligner~\cite{McAuliffe2017}\footnote{\url{https://github.com/MontrealCorpusTools/Montreal-Forced-Aligner}}.

\subsection{Manual Annotation Details}\label{app:man}

\begin{figure*}[htbp]
    \centering
    \includegraphics[width=1\linewidth]{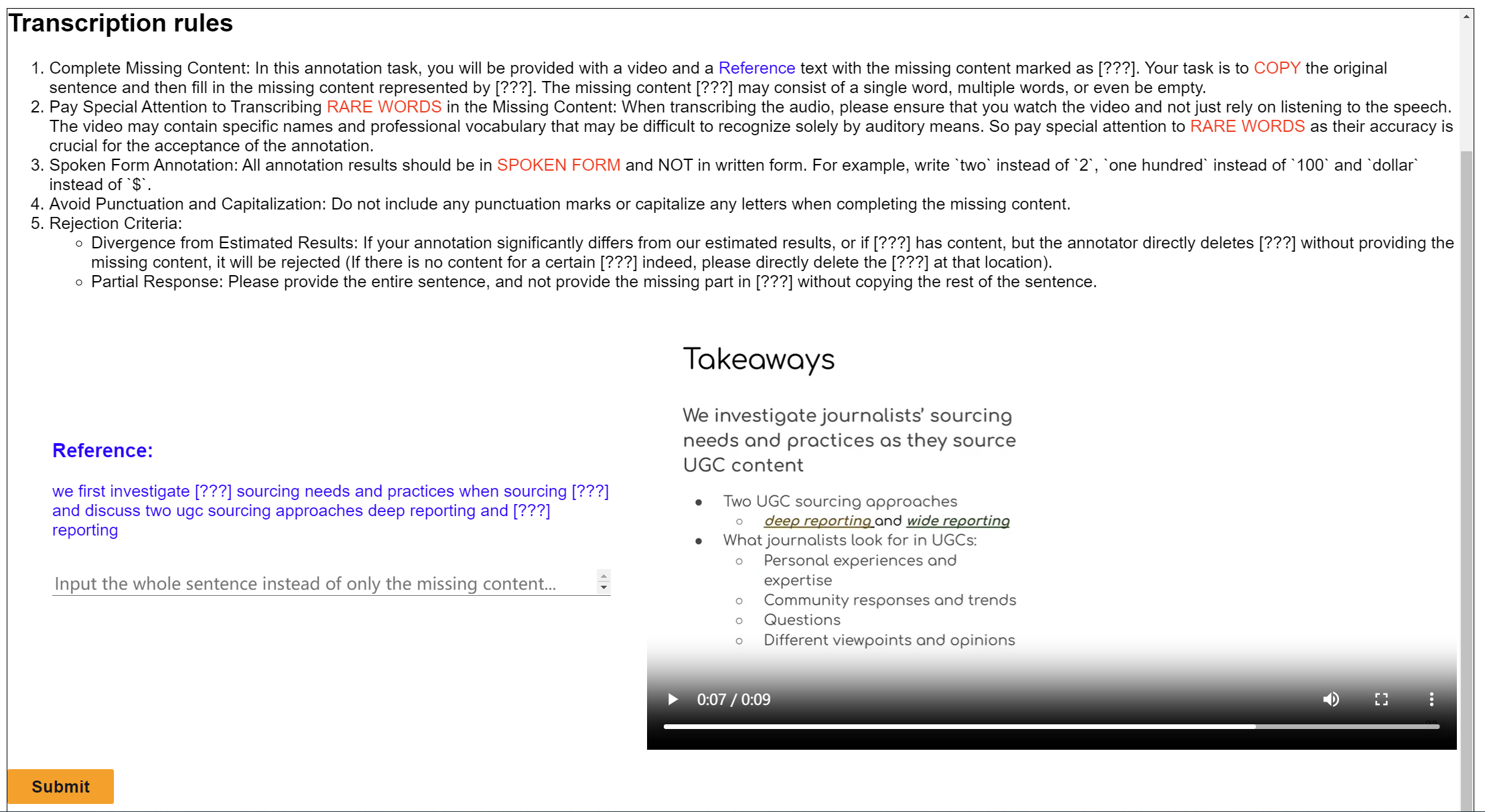}
    \caption{Annotation interface for manual filling in constructing speech transcription.}
    \label{fig:mturk_fill}
\end{figure*}

\begin{figure*}[htbp]
    \centering
    \includegraphics[width=1\linewidth]{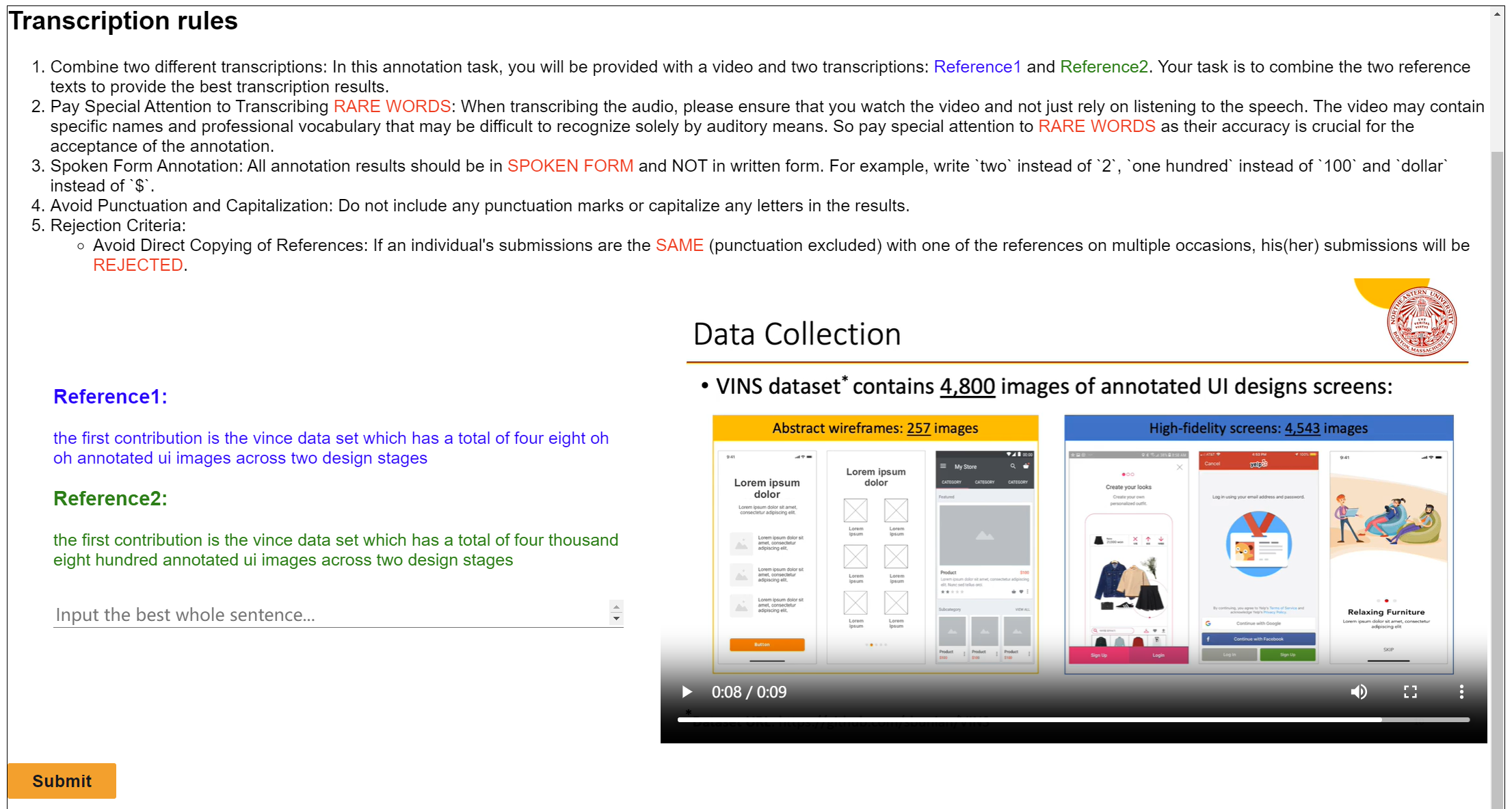}
    \caption{Annotation interface for manual combination in constructing speech transcription.}
    \label{fig:mturk_combine}
\end{figure*}

The master-granted annotators are assigned to complete manual filling and combination in creating speech transcription. 
In these two tasks, each worker receives 0.06 USD for each finished sample, respectively. 
The annotation interfaces shown in Figure~\ref{fig:mturk_fill} and Figure~\ref{fig:mturk_combine} describe the rules as well as the rejection cases of manual filling and combination in detail, respectively. The failed samples are passed to other workers for annotation.

\begin{figure*}[htbp]
    \centering
    \includegraphics[width=1\linewidth]{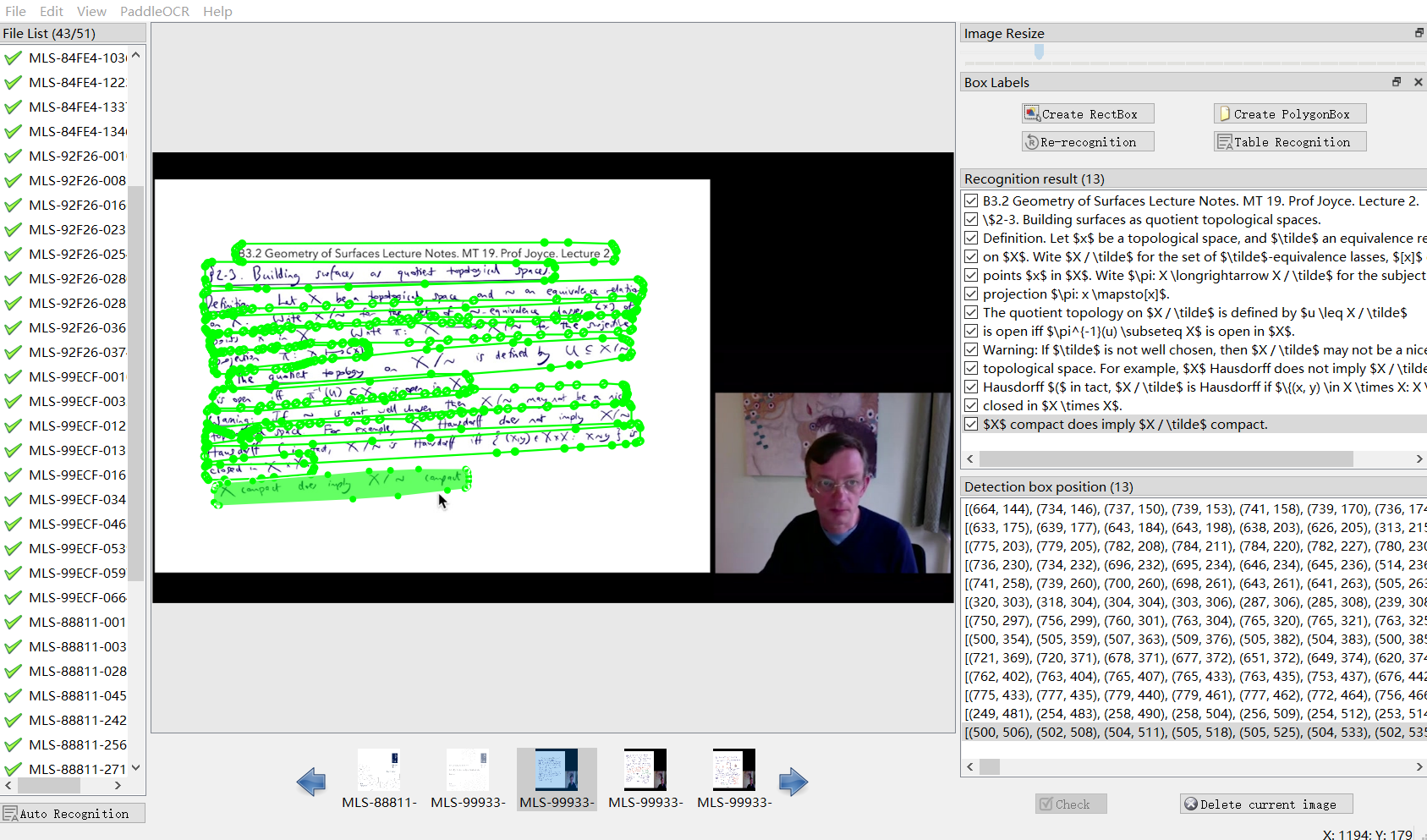}
    \caption{Annotation interface for correcting OCR results.}
    \label{fig:ppocrlabel}
\end{figure*}

The experienced researchers in our team are assigned to annotate the OCR data, and the annotation of each sample includes slide deduplication and OCR content correction. 
Each worker received the appropriate remuneration.
The interface for OCR content correction is shown in Figure~\ref{fig:ppocrlabel}, where PPOCRLabel\footnote{\url{https://github.com/PaddlePaddle/PaddleOCR/tree/release/2.7/PPOCRLabel}} is adopted. To control the quality of labelling, we sampled the results of 9 annotators individually and checked whether 120 of the annotated images had a high number of missing or incorrect items, with no more than 10 wrong items.

\begin{figure*}[htbp]
    \centering
    \includegraphics[width=1\linewidth]{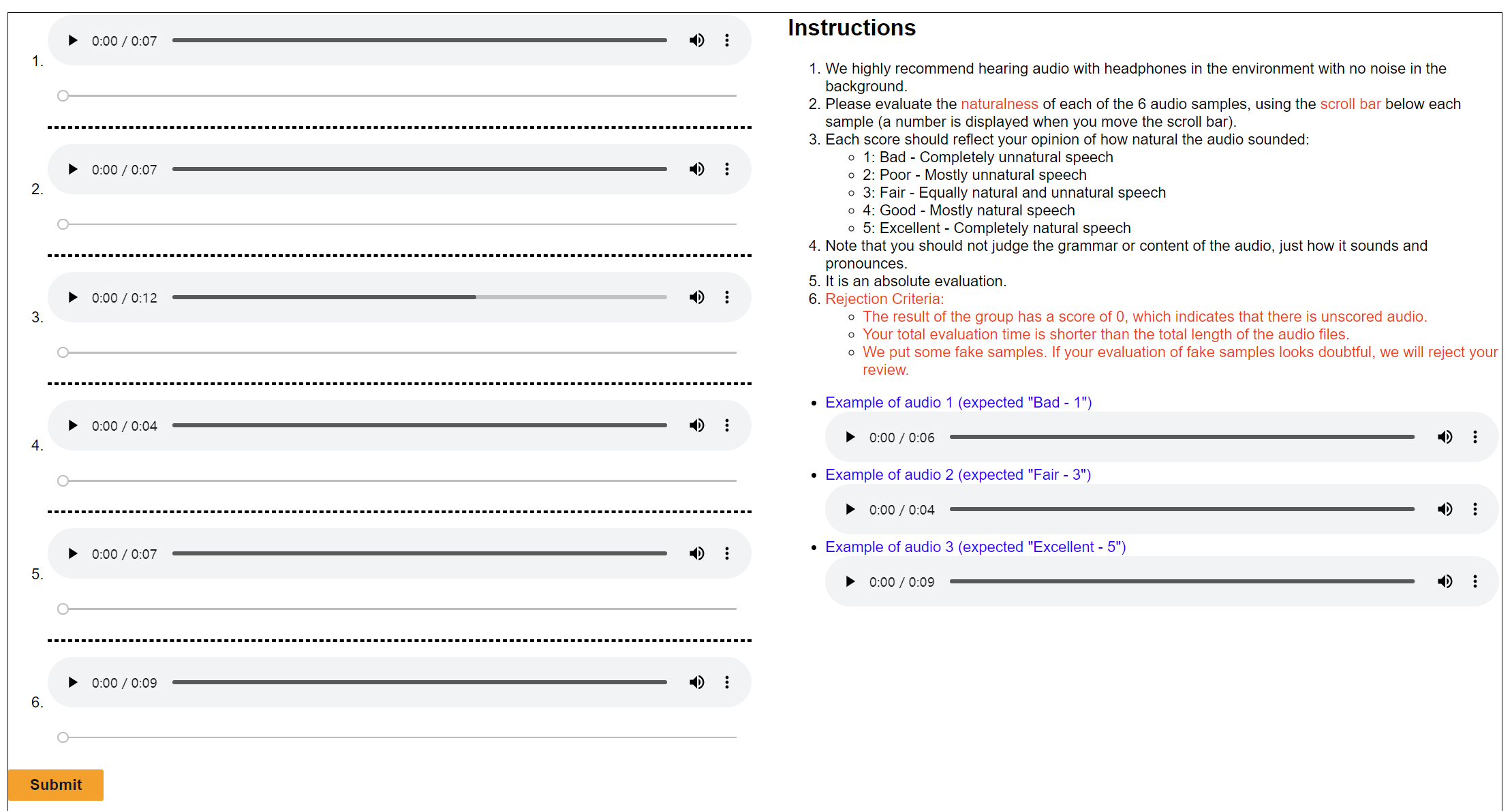}
    \caption{Annotation interface for MOS scoring in evaluating TTS.}
    \label{fig:mturk_mos}
\end{figure*}

The master-granted annotators are assigned to complete the MOS scoring in the TTS task evaluation. A batch of 6 TTS results from the same audio clip (with the addition of a trap sample used to detect attentiveness). 
Each worker receives 0.2 USD for each completed batch respectively.
Similarly, the annotation interface shown in Figure~\ref{fig:mturk_mos} details the rules as well as the rejection cases, which is inspired by \citet{Choi2023}. The failed samples will be passed to other workers for annotation.

\begin{figure*}[htbp]
    \centering
    \includegraphics[width=1\linewidth]{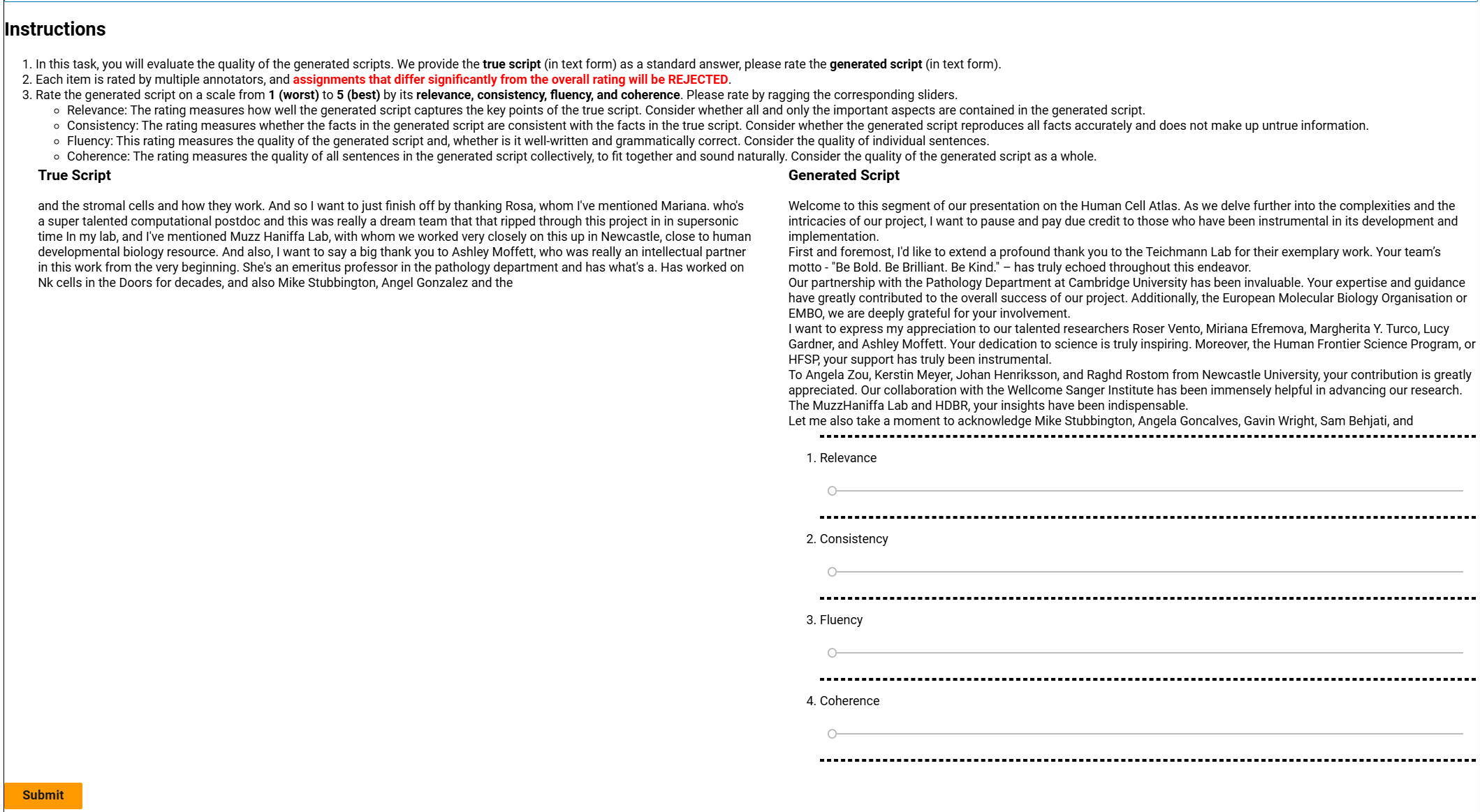}
    \caption{Annotation interface for manual scoring in Slide $\rightarrow$ Script task.}
    \label{fig:mturk_slide2speech}
\end{figure*}

\begin{figure*}[htbp]
    \centering
    \includegraphics[width=1\linewidth]{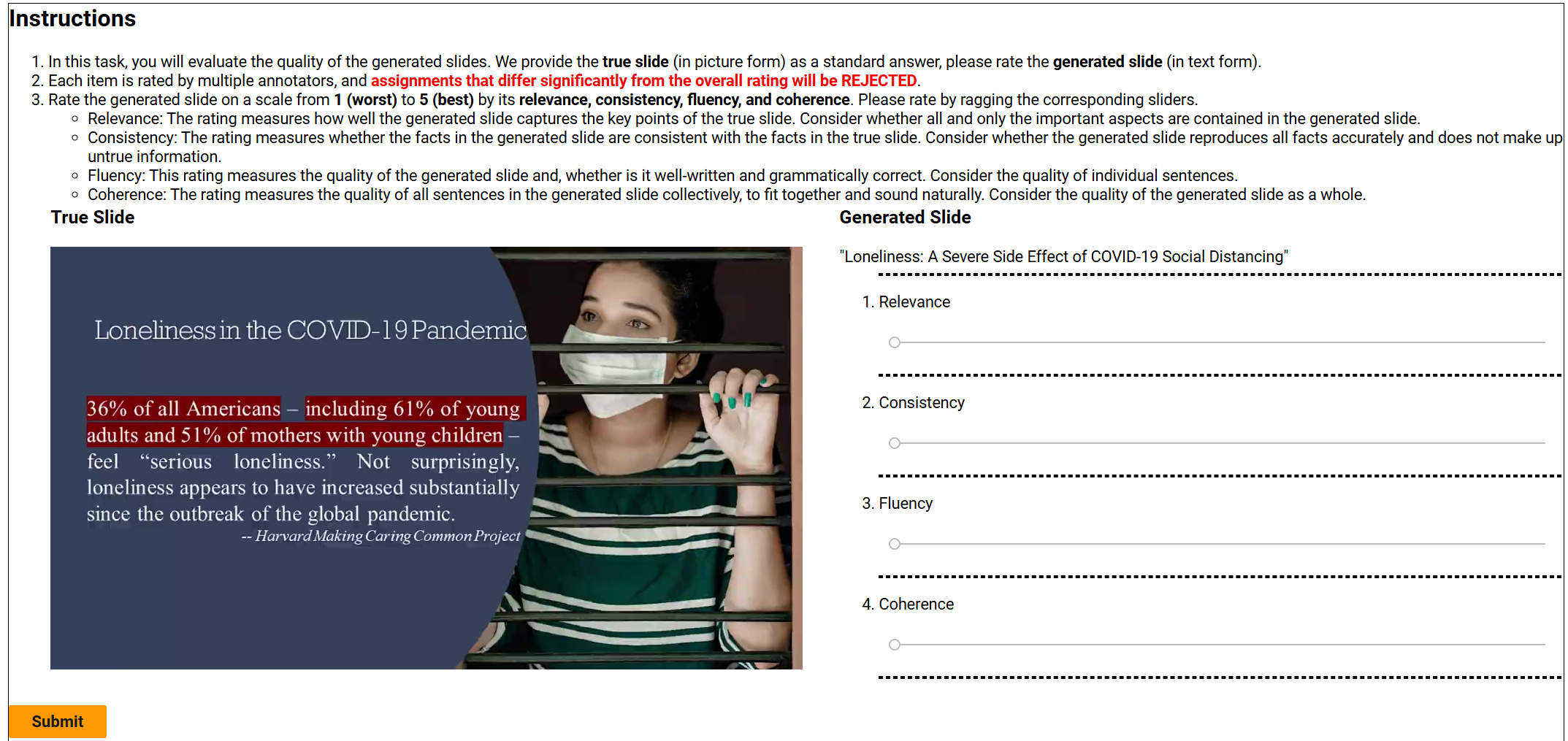}
    \caption{Annotation interface for manual scoring in Script $\rightarrow$ Slide task.}
    \label{fig:mturk_speech2slide}
\end{figure*}

As for the manual scoring of SSG tasks, we posted the annotation task on Amazon Mechanical Turk initially but found that the quality was unsatisfactory due to the complexity of the scripts and speech texts. Therefore, three experienced researchers were assigned to carry out the annotation task. Specifically, annotators need to rate the generated slides/scripts on multiple dimensions, including coherence, consistency, fluency and relevance, as referenced in SummEval~\cite{fabbri2021summeval}. Each sample is scored on a scale from 1 (worst) to 5 (best) by 3 annotators. Note that we sampled 100 items due to cost and multiple models. The annotation interfaces are shown in Figure~\ref{fig:mturk_slide2speech} and Figure~\ref{fig:mturk_speech2slide}.

\section{Experiment Details}

\subsection{ASR Benchmark Details}\label{app:asr}

We construct the Attention-based Encoder-Decoder (AED) with the help of the well-known ESPnet~\cite{Watanabe2018} toolkit. We use Conformer~\cite{Gulati2020} as the encoder, which consists of 12 blocks, with hidden linear units $d^{\text{ff}}=2048$ and attention output size $d^{\text{att}}=512$. Attention head $H$ is set to 8 and the front CNN kernel size is 31. As for the decoder, we use the Transformer with 6 blocks ($d^{\text{ff}}=2048$,$H=8$). We use BPE as our recognition unit, which consists of 1k tokens obtained by SentencePiece~\cite{Kudo2018}. In addition to the attention training objectives, We use a certain degree of CTC loss function. CTC weight $\lambda$ is 0.3 during training, while 0.2 for infering. SpecAugment~\cite{Park2019} is applied with time mask width $T=40$ and frequency mask with $F=30$. Our model is trained on four 24GB 3090 RTX GPUs for 60 epochs. The top 10 checkpoints after one epoch training are preserved for model averaging. 

For the RNN-Transducer (RNN-T), we also adopt Conformer as the encoder, with an intermediate joint network of 320 dimensions. The Conformer has 15 blocks, 1024 hidden linear units, 4 attention heads, and 256 attention output dimensions. After that, a single layer of LSTM is performed.

\subsection{CASR Benchmark Details}\label{app:casr}

Our CASR task is performed under two paradigms. One paradigm is to form the biasing list for an utterance using target biasing words together with distractors that are randomly selected from all rare words, as a simulation of the real-world scenario \cite{Aleksic2015,Zhao2019,Pundak2018,Le2021a}. Our experiment uses 1000 distractors to form the biasing list for each utterance, and rare words are defined by removing the top 5k most frequent words from the vocabulary.
Another paradigm is audio-visual CASR~\cite{Sun2022} which extracts the biasing list from visual information, such as OCR annotation. 

 As for TCPGen, we use a 6-layer GCN~\cite{Kipf2017} for future information representation. Following the method proposed in ~\cite{Le2021}, the biasing lists are extracted by identifying words in the speech transcription of each utterance that also appears in the total rare words list, followed by the addition of plenty of distractors to simulate a real-life setting. The length of the biasing list is set to 500 during training and 1000 during inferencing. Model training lasts for 70 epochs, and the best 10 of the checkpoints are saved for model averaging. Our models are trained on 24GB 3090 RTX GPUs.

Before model training, the rare word selection is first performed. Our speech corpus includes approximately 43k different words in total. We conduct a word frequency analysis on the training set, identifying words with a frequency ranking exceeding 5k as members of the total rare word set. The number of our rare word list is about 38k.  

\subsection{Spontaneous TTS Benchmark Details}\label{app:tts}

We followed the settings recommended by the architecture authors, and their experiments were trained on a subset of GigaSpeech~\cite{Chen2021}, which is a speech corpus containing transcribed audio from audiobooks, Podcasts, and YouTube. We use the training set of the M$^3$AV for MQTTS model training and arbitrarily extract a sufficient number of samples from the Dev and Test Set for speech synthesis to verify the effectiveness of our model. We adopt HiFi-GAN~\cite{Kong2020} as the backbone architecture for the quantizer and discriminator, which we trained for 600k steps. After that, a Transformer is trained to autoregressively generate the code representation based on the past code sequence. The last checkpoints from both steps are preserved. It is worth noting that MQTTS training may take a long time. Our models are trained on 24GB 3090 RTX GPUs. 

\subsection{Slide and Script Generation Benchmark Details}\label{app:slide}

For the selection of related paper sentences, we adopt the DistilBERT~\cite{Sanh2019} model in the repository of sentence-transformers~\cite{Reimers2019}~\footnote{\url{https://huggingface.co/sentence-transformers/msmarco-distilbert-base-v4}} to calculate the similarity between the script text and each paper sentence. Then we select the top 3 candidates with similarity greater than 0.5 as the selection results.
In the training phase of benchmark systems, we use LoRA~\cite{Hu2022} to fine-tune LLaMA-2~\cite{Touvron2023}. The optimizer we use is AdamW~\cite{Loshchilov2018}. The learning rate curve is 1 epoch of warm-up and then drops with a peak learning rate of $2\times 10^{-4}$. The total number of training epochs is 3. Our models are trained on 24GB 3090 GPUs and 48GB A6000 GPUs. Fine-tuned InstructBLIP~\cite{Dai2023} is another baseline. Specifically, we fine-tune the officially released checkpoint for 20 epochs with the learning rate increasing linearly from $1\times 10^{-7}$ to $5\times 10^{-5}$ during the first 1000 steps and dropping to 0 as a cosine function. The model is trained on 80GB A100 GPUs with AdamW~\cite{Loshchilov2018} as the optimizer. The best checkpoints are chosen according to the performance of the development set. 
As for the generation settings, we use nucleus sampling~\cite{Holtzman2020} with a $\text{top}_\text{p}$ of 0.9. The maximum number of tokens generated is 300. These generation settings are consistent across LLaMA-2, InstructBLIP, GPT-4 and GPT-4V. 

\subsection{Detailed Manual Scoring Results on SSG Tasks}\label{ssg_detail_score}

We show the detailed scoring results in Table~\ref{tab:ssg_score_1} and Table~\ref{tab:ssg_score_2}, which include the coherence (Coh.), consistency (Cons.), fluency (Flu.), relevance (Rel.) and average value (Avg.) between the hypothesis and its reference text.

\begin{table}[t]
\renewcommand{\arraystretch}{0.9}
\caption{Detailed manual scoring results on SSG tasks.}
\vspace{-0.2cm}
\centering
\resizebox{1\linewidth}{!}{
\begin{tabular}{l c c c c c}
\toprule
\bf Model & \bf Coh. ($\uparrow$) & \bf Cons. ($\uparrow$) & \bf Flu. ($\uparrow$) & \bf Rel. ($\uparrow$) & \bf Avg. ($\uparrow$)\\
\midrule 
\multicolumn{6}{c} {\emph{Slide $\longrightarrow$ Script}}\\
\midrule
LLaMA-2$_{\textsc{7B}}$+OCR & 3.2 & 3.0 & 4.2 & 4.1 & 3.6 \\
LLaMA-2$_{\textsc{13B}}$+OCR & 3.7 & 3.4 & 4.1 & 4.3 & 3.9 \\
GPT-4+OCR & 4.0 & 4.4 & \bf 5.0 & \bf 5.0 & 4.6 \\
\midrule
InstructBLIP$_{\textsc{7B}}$ & 1.6 & 2.0 & 2.6 & 3.2 & 2.4 \\
InstructBLIP$_{\textsc{13B}}$ & 2.0 & 2.1 & 2.6 & 3.3 & 2.5 \\
GPT-4V & \bf 4.2 & \bf 4.5 & \bf 5.0 & \bf 5.0 & \bf 4.7 \\
\midrule
\multicolumn{6}{c} {\emph{Script $\longrightarrow$ Slide}}\\
\midrule
LLaMA-2$_{\textsc{7B}}$ & 3.1 & 3.4 & 3.4 & 3.3 & 3.3 \\
LLaMA-2$_{\textsc{13B}}$ & 3.5 & 3.6 & 3.9 & 4.0 & 3.8 \\
GPT-4 & \bf 4.6 & \bf 4.4 & \bf 5.0 & \bf 5.0 & \bf 4.8 \\
\bottomrule
\end{tabular}
}
\label{tab:ssg_score_1}
\end{table}

\begin{table}[t]
\renewcommand{\arraystretch}{0.8}
\caption{Detailed manual scoring results on SSG tasks when paper information is introduced.}
\vspace{-0.2cm}
\centering
\resizebox{1\linewidth}{!}{
\begin{tabular}{l c c c c c}
\toprule
\bf Range & \bf Coh. ($\uparrow$) & \bf Cons. ($\uparrow$) & \bf Flu. ($\uparrow$) & \bf Rel. ($\uparrow$) & \bf Avg. ($\uparrow$)\\
\midrule 
\multicolumn{6}{c} {\emph{Slide $\longrightarrow$ Script}}\\
\midrule
Subset & 3.4 & 3.2 & \bf 4.2 & \bf 4.1 & 3.7 \\
~~ + Paper & \bf 4.1 & \bf 4.2 & 4.1 & \bf 4.1 & \bf 4.1 \\
\midrule
\multicolumn{6}{c} {\emph{Script $\longrightarrow$ Slide}}\\
\midrule
Subset & 3.2 & 3.4 & 3.3 & 3.3 & 3.3 \\
~~ + Paper & \bf 3.6 & \bf 3.7 & \bf 3.5 & \bf 3.4 & \bf 3.6 \\
\bottomrule
\end{tabular}
}
\label{tab:ssg_score_2}
\vspace{-0.4cm}
\end{table}

\section{Qualitative Analyses of SSG Benchmark Systems}\label{sec:cases}

\begin{figure}[htbp]
    \centering
    \includegraphics[width=1\linewidth]{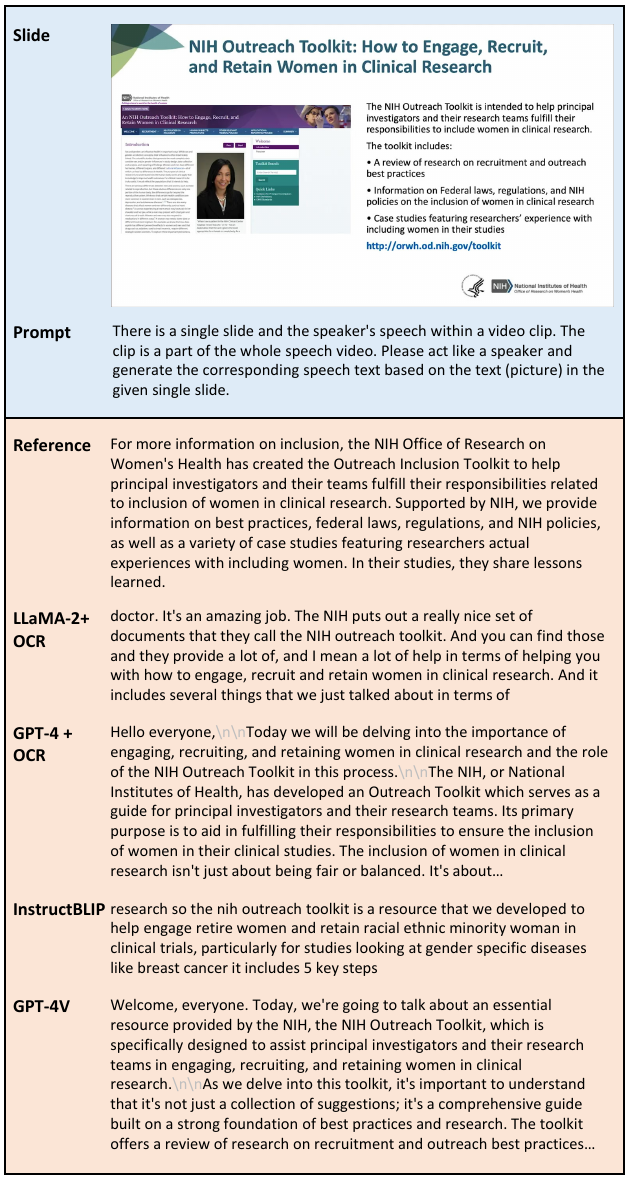}
    \caption{``Slide$\rightarrow$Script'' Case I where scripts generated by benchmark systems are compared.}
    \label{fig:case1}
\end{figure}

\begin{figure}[htbp]
    \centering
    \includegraphics[width=1\linewidth]{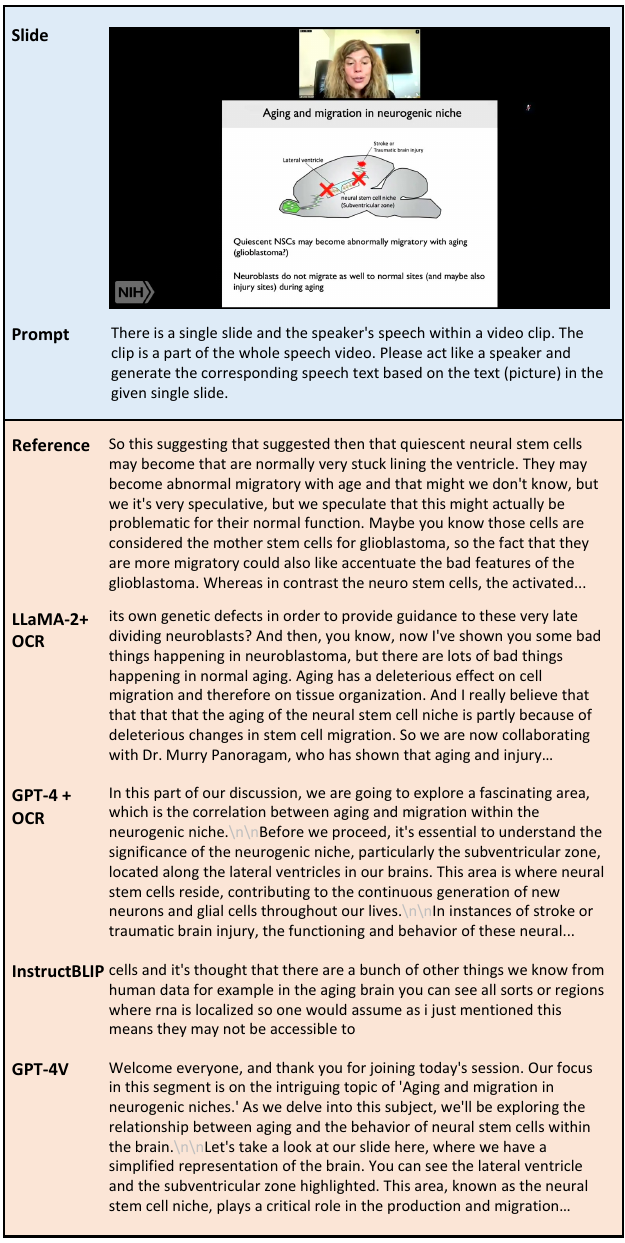}
    \caption{``Slide$\rightarrow$Script'' Case II where scripts generated by benchmark systems are compared.}
    \label{fig:case2}
\end{figure}

The comparisons of the benchmark systems in ``Slide$\rightarrow$Script'' are shown in Figure~\ref{fig:case1} and Figure~\ref{fig:case2}. LLaMA-2 and InstructBLIP below refer to models of size 7B. We can find that the scripts generated by GPT-4+OCR and GPT-4V are highly qualified and roughly match the references. The results of LLaMA-2+OCR are acceptable, although slightly weaker than the previous two. All three can develop detailed descriptions around topics in slides, such as the ``NIH Outreach Toolkit'' in Figure~\ref{fig:case1} and ``Aging and Migration'' in Figure~\ref{fig:case2}. InstructBLIP, on the other hand, fails to find the right theme, as shown in Figure~\ref{fig:case1} and Figure~\ref{fig:case2}, suggesting that today's open-source LMM models struggle in handling text in slides.

\begin{figure}[htbp]
    \centering
    \includegraphics[width=1\linewidth]{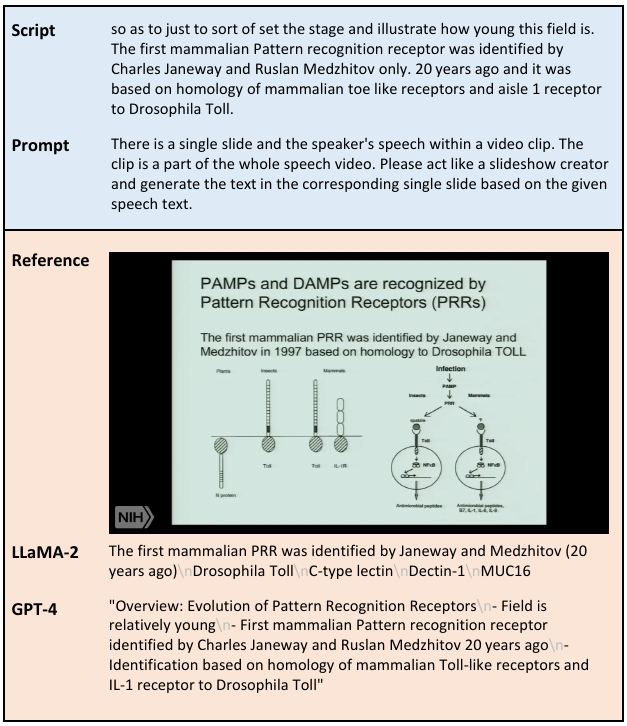}
    \caption{``Script$\rightarrow$Slide'' Case I where slides generated by benchmark systems are compared.}
    \label{fig:case4}
\end{figure}

\begin{figure}[htbp]
    \centering
    \includegraphics[width=1\linewidth]{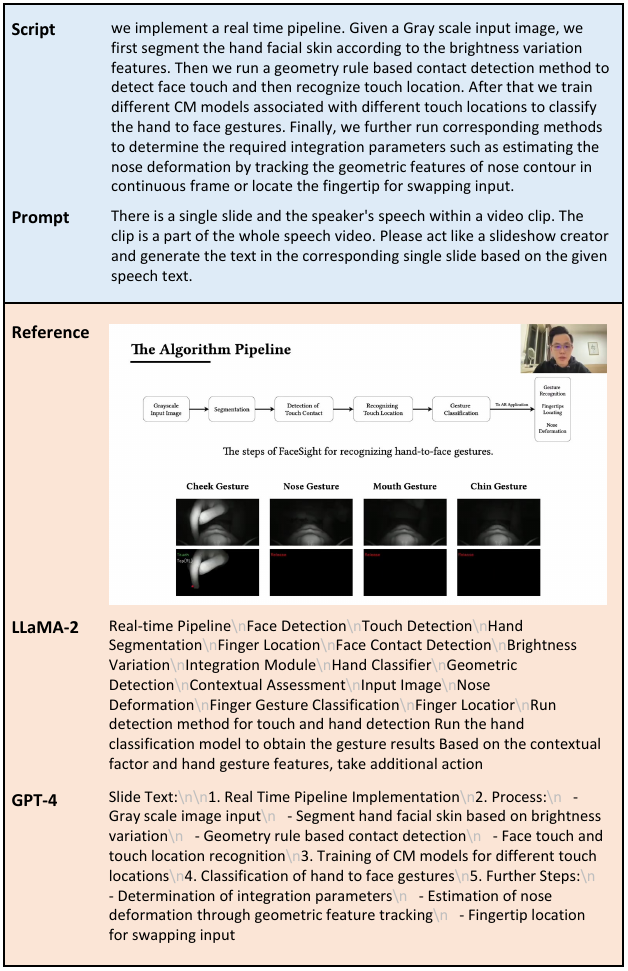}
    \caption{``Script$\rightarrow$Slide'' Case II where slides generated by benchmark systems are compared.}
    \label{fig:case5}
\end{figure}

The comparisons of the benchmark systems in ``Script$\rightarrow$Slide'' are shown in Figure~\ref{fig:case4} and Figure~\ref{fig:case5}. The slides generated by GPT-4 are closer to references than those generated by LLaMA-2, and the contents are more accurate and condensed. For example, GPT-4 presents ``Drosophila ToLL'' in Figure~\ref{fig:case4} and ``Pipeline'' in Figure~\ref{fig:case5}, while LLaMA-2 does not. We can conclude that GPT-4 has a better understanding of complex academic knowledge than LLaMA-2.

\begin{figure}[htbp]
    \centering
    \includegraphics[width=1\linewidth]{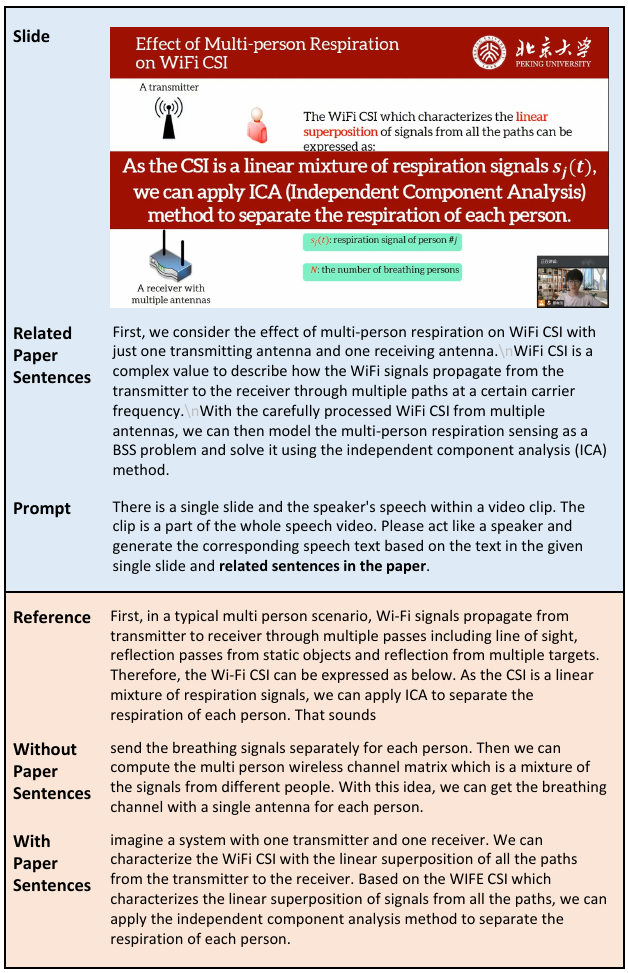}
    \caption{``Slide$\rightarrow$Script'' Case III where paper sentences are provided as external knowledge.}
    \label{fig:case3}
\end{figure}

\begin{figure}[htbp]
    \centering
    \includegraphics[width=1\linewidth]{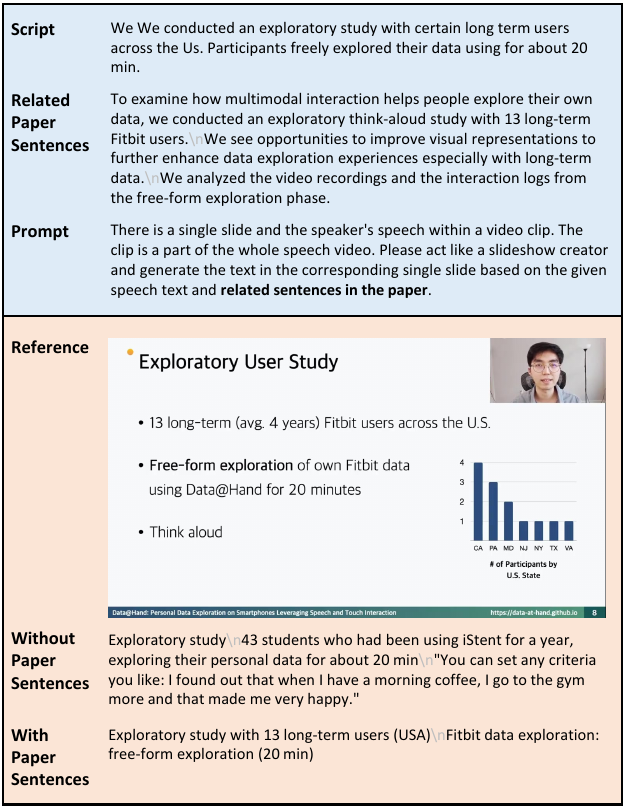}
    \caption{``Script$\rightarrow$Slide'' Case III where paper sentences are provided as external knowledge.}
    \label{fig:case6}
\end{figure}

Figure~\ref{fig:case3} and Figure~\ref{fig:case6} show the improvement in the quality of the generated scripts and slides brought by external knowledge. The addition of external knowledge enables AI models to produce necessary details, such as ``WiFi CSI'' in Figure~\ref{fig:case3} and ``13 long-term users'' in Figure~\ref{fig:case6}. It shows that the introduction of external knowledge leads to more comprehensive information captured by AI models.

\end{document}